\documentclass[sigconf, screen, nonacm]{acmart}
\usepackage{enumitem}
\usepackage{subcaption}
\usepackage{microtype}
\usepackage{diagbox}
\usepackage{amsmath}
\newcommand{\rev}[1]{{#1}}
\AtBeginDocument{%
  }

\copyrightyear{2026}
\acmYear{2026}
\setcctype{by}
\acmConference[HRI '26]{Proceedings of the 21st ACM/IEEE International Conference on Human-Robot Interaction}{March 16--19, 2026}{Edinburgh, Scotland, UK}
\acmBooktitle{Proceedings of the 21st ACM/IEEE International Conference on Human-Robot Interaction (HRI '26), March 16--19, 2026, Edinburgh, Scotland, UK}
\acmDOI{10.1145/3757279.3788670}
\acmISBN{979-8-4007-2128-1/2026/03}
\received{2025-09-30}
\received[accepted]{2025-12-23}

\begin{document}

\title{Towards Intelligible Human-Robot Interaction: An Active Inference Approach to Occluded Pedestrian Scenarios}

\author{Kai Chen}
  \affiliation{
  \institution{Tongji University}
  \city{Shanghai}
  \country{China}}
  \email{14sdck@tongji.edu.cn}
  \orcid{0000-0001-7211-5112}

\author{Yuyao Huang}
\affiliation{%
  \institution{Tongji University}
  \city{Shanghai}
  \country{China}}
  \email{huangyuyao@tongji.edu.cn}
  \orcid{0000-0001-6723-5582}

\author{Guang Chen}
  \affiliation{%
  \institution{Tongji University}
  \city{Shanghai}
  \country{China}}
  \email{guangchen@tongji.edu.cn}
  \orcid{0000-0002-7416-592X}

\begin{abstract}

The sudden appearance of occluded pedestrians presents a critical safety challenge in autonomous driving. Conventional rule-based or purely data-driven approaches struggle with the inherent high uncertainty of these long-tail scenarios. To tackle this challenge, we propose a novel framework grounded in Active Inference, which endows the agent with a human-like, belief-driven mechanism. Our framework leverages a Rao-Blackwellized Particle Filter (RBPF) to efficiently estimate the pedestrian's hybrid state. To emulate human-like cognitive processes under uncertainty, we introduce a Conditional Belief Reset mechanism and a Hypothesis Injection technique to explicitly model beliefs about the pedestrian's multiple latent intentions. Planning is achieved via a Cross-Entropy Method (CEM) enhanced Model Predictive Path Integral (MPPI) controller, which synergizes the efficient, iterative search of CEM with the inherent robustness of MPPI. Simulation experiments demonstrate that our approach significantly reduces the collision rate compared to reactive, rule-based, and reinforcement learning (RL) baselines, while also exhibiting explainable and human-like driving behavior that reflects the agent's internal belief state.

\end{abstract}




\maketitle

\section{Introduction}

Achieving safe interaction in complex driving scenarios remains a significant challenge, particularly in highly uncertain situations, such as occlusions. Although data-driven methods, including reinforcement learning, imitation learning, and large language models, have shown considerable success~\cite{zhao2024dl_review, huang2025vlm_rl, zhou2024vlm_survey, zhou2024dl_survey}, they often encounter substantial hurdles. These include the scarcity of data for safety-critical corner cases~\cite{kiran2021deep, chen2024end} and a lack of interpretability~\cite{atakishiyev2024explainable}. Consequently, several mechanism-inspired approaches~\cite{johan2024surprise, albrecht2021goal, jiang2022personalized} have been proposed. Among these, the free energy principle~\cite{friston2010free_energy} and its process theory, active inference~\cite{active_inference_book, friston2017process_theory}, stand out as theoretically rigorous and promising. They offer a biologically inspired framework with applications spanning robotics~\cite{pio2016aif_robot}, autonomous driving~\cite{engstrom2024resolving, schumann2025active_collision, wei2025car_following, wei2024navigation}, and human-computer interaction (HCI)~\cite{stein2024towards_interaction}. Recently, researchers~\cite{lu2025human_like_driving} have characterized the exploratory applications of active inference in traffic occlusion scenarios~\cite{engstrom2024resolving} as a form of \textit{mechanism-inspired human-like driving}, positioning it as a paradigm for autonomous decision-making. Furthermore, a recent review~\cite{murray2024aif_hci} highlights active inference as a suitable framework for modeling the human-computer interaction loop.

\begin{figure}[htbp]
  \centering
  \begin{subfigure}[b]{0.8\linewidth}
    \includegraphics[width=\linewidth, alt={A reactive agent failing to slow down for an occluded pedestrian and colliding.}]{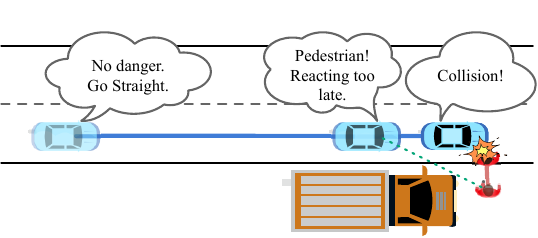}
    \caption{}
    \Description{}
    \label{fig:reactive}
  \end{subfigure}
  
  \begin{subfigure}[b]{0.8\linewidth}
    \includegraphics[width=\linewidth, alt={A proactive agent slowing down and swerving to safely avoid the same pedestrian.}]{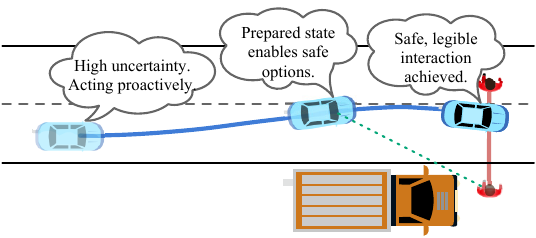}
    \caption{}
    \Description{}
    \label{fig:active}
  \end{subfigure}
  \caption{Reactive vs. Proactive Driving in a Critical Occluded Pedestrian Scenario. (a) A reactive agent ignores the uncertainty from occlusion, fails to slow down, and collides. (b) Our proactive agent reasons about the latent hazard, preemptively reducing speed to ensure \rev{a safe interaction with interpretable behaviors.}}
  \Description{Two diagrams comparing a reactive driving agent to a proactive one. The top diagram shows a reactive agent failing to slow down for an occluded pedestrian and colliding. The bottom diagram shows a proactive agent slowing down and swerving to safely avoid the same pedestrian.}
  \label{fig:intro}
\end{figure}

Motivated by this background, we propose an interactive framework based on active inference to address the occluded pedestrian scenario. From the perspective of human-like driving~\cite{lu2025human_like_driving}, our approach is a learning-free, mechanism-inspired method. Within the theoretical framework of active inference for HCI~\cite{murray2024aif_hci}, it is a variant of a reflective model, where the agent internally maintains a generative model of the potential pedestrian and leverages this model for planning and action, resulting in the proactive, human-like driving behavior illustrated in Figure~\ref{fig:intro}.

\rev{Our framework implements active inference via a continuous perception-action loop governed by the single principle of free energy minimization. To address the specific challenges of occlusion, we integrate two cognitively-inspired mechanisms into this cycle. First, we enhance the \textit{Belief Updating} process with a novel \textit{Conditional Reset} mechanism. This prevents the agent's belief in a potential hazard from premature decay due to occlusion, thereby mirroring the vigilance of a human driver. Concurrently, we augment the \textit{Planning as Inference} process with \textit{Hypothesis Injection}. This mechanism emulates human-like \textit{what-if} reasoning, forcing the planner to proactively prepare for worst-case pedestrian behaviors, ensuring safety despite uncertainty.}

Our main contributions are threefold:
\begin{itemize}[leftmargin=14pt] 
    \item We propose a novel active inference framework to address the occluded pedestrian scenario, formalizing the interaction as a POMDP with a rich set of latent pedestrian states.
    \item We introduce two novel, cognitively-inspired mechanisms to the active inference framework: \textit{Conditional Reset} to ensure belief persistence under uncertainty, and \textit{Hypothesis Injection} to enable proactive planning through counterfactual reasoning.
    \item We validate our active inference interaction framework extensively in a challenging simulation, showing it produces \rev{explainable}, human-like driving behaviors and outperforms baseline methods in safety-critical metrics.
\end{itemize}

\section{Related Work}
\label{sec:related_work}

Active inference is emerging as a powerful, first-principles framework for modeling human-like driving behavior, offering an interpretable alternative to black-box models. Our work builds directly on the foundational framework of \cite{engstrom2024resolving}, who first modeled adaptive driving as a trade-off between pragmatic (goal-seeking) and epistemic (information-seeking) value. We extend their work by specializing it for the occluded pedestrian scenario, expanding the pedestrian's generative model and introducing two novel mechanisms---\textit{Conditional Reset} and \textit{Hypothesis Injection}---to emulate human-like belief maintenance and proactive reasoning. \rev{This probabilistic cognitive modeling differs fundamentally from game-theoretic approaches like \cite{zhang2021safe}, which handle uncertainty through set-based reachability analysis to provide worst-case safety guarantees.} Similarly, \cite{schumann2025active_collision} propose a descriptive cognitive model to reproduce empirical human collision avoidance data. Our work differs in its prescriptive goal of engineering a safe and \rev{interpretable} autonomous agent, rather than explaining existing human behavior.
In contrast to mechanism-based approaches, a significant body of work focuses on learning the components of the active inference agent or the planning policy directly from data. These methods include learning driver preferences from demonstrations via inverse reinforcement learning~\cite{wei2025car_following}; using imitation learning to initialize a Dynamic Bayesian Network updated online~\cite{nozari2024modeling, nozari2022active}; and leveraging deep generative models to learn a forward model of the environment from high-dimensional sensory inputs~\cite{catal2020deep, delavari2024towards, huang2024diffusion, wei2024switching}. \rev{Specifically addressing interaction and occlusion, \cite{hu2023deception} utilize adversarial reinforcement learning to approximate solutions for belief-space safety games, while \cite{packer2023anyone} employ flow-based generative models trained on offline datasets to perform contingency planning for unobserved agents.} Although data-driven methods excel at capturing driving behaviors present in a training set, our mechanism-based approach is designed for robustness in safety-critical, long-tail scenarios that are often absent from such datasets.

\section{Methods}

\subsection{Preliminaries: Active Inference Framework}

\rev{Our work employs Active Inference~\cite{active_inference_book}, a first-principles framework from computational neuroscience, to model the complex cognitive processes in vehicle-pedestrian interaction. Active Inference posits that an agent minimizes the discrepancy between sensory observations and predictions from an internal generative model by minimizing a single objective: free energy.}

\rev{This framework unifies perception and action. \textbf{Perception} is modeled as inferring the hidden causes of sensory signals by minimizing \textit{Variational Free Energy (VFE)}. This ensures that the agent's belief becomes the best possible approximation of the true posterior distribution over hidden states given the available data. \textbf{Action} is selected to minimize \textit{Expected Free Energy (EFE)} over a future horizon. EFE naturally balances \textit{pragmatic value} (goal-seeking behavior) and \textit{epistemic value} (information-seeking behavior).}

In our proposed framework, minimizing VFE drives the \textit{Belief Updating} process (incorporating new observations), while minimizing EFE drives the \textit{Planning} process (resolving uncertainty and avoiding collisions). Details of VFE and EFE are provided in Appendix~\ref{app:A}.

\subsection{Problem Definition}
\label{subsec:problem}

We formalize the dynamic interaction between an ego-vehicle and a potential pedestrian near an occluder as a sequential decision-making problem under uncertainty extended from \cite{engstrom2024resolving}. The objective of the ego-vehicle is to derive a policy that safely and efficiently resolves this interaction, despite profound uncertainty regarding the pedestrian's existence, state, and intentions. This interactive challenge is modeled as a Partially Observable Markov Decision Process (POMDP)~\cite{smallwood1973pomdp}.

\subsubsection{System State}
We use subscripts $e$, $p$, and $c$ to denote the ego-vehicle, pedestrian and collision, respectively. The complete state of the environment at time $t$ consists of the ego-vehicle, the pedestrian, and the static surroundings.

\subparagraph{\textbf{Ego-Vehicle State}} The kinematic state of the ego-vehicle is defined as $s_{e,t} = (x_{e, t}, v_{e, t}, a_{e, t}) \in \mathbb{R}^6$, 
representing its position, velocity, and acceleration. The vehicle has fixed length $L_e$ and width $W_e$.

\subparagraph{\textbf{Static Environment}} The environment contains a static occluding object $\mathcal{M}_\mathrm{occ}$ with dimensions $L_\mathrm{occ}$ and $W_\mathrm{occ}$, centered at $x_\mathrm{occ}$.

\rev{\subparagraph{\textbf{Hidden Pedestrian State}} The pedestrian's state is hidden and consists of: an existential state $z_p \in \{0, 1\}$ (indicating presence or absence); a kinematic state $s_{p,t} = (x_{p, t}, v_{p, t}, a_{p, t}) \in \mathbb{R}^6$ (assuming lateral movement perpendicular to the lane); and a set of latent behavioral parameters $\theta_p$. These parameters $\theta_p$ govern the pedestrian's specific behavioral logic and motion characteristics (e.g., aggressiveness) and include the motion activation distance relative to the ego-vehicle ($d_\mathrm{act}$), as detailed in Appendix~\ref{app:B}.}

\subsubsection{Action and Observation}
\label{sec:observation}
The ego-vehicle's action $a_{e,t} \in \mathbb{R}^2$ consists of longitudinal and lateral acceleration. The observation $o_t = (o_{p,t}, \boldsymbol{x}_{p, t}, o_{c, t})$ comprises a binary visibility status $o_{p,t}$, the pedestrian's measured position $\boldsymbol{x}_{p, t}$ and a collision status $o_{c, t}$.

\subsection{Active Inference Interaction Framework}

\begin{figure*}[ht]
  \centering
  \includegraphics[width=0.95\linewidth, alt={This is a comprehensive multi-part diagram illustrating an autonomous interaction framework, organized into three main sections: a top row of high-level flowcharts, a bottom row of concrete scenario illustrations, and a legend at the very bottom. The top section is divided into three interconnected parts. On the top-left, a light green box titled "Planning as Inference" shows a "CEM Iteration" loop. This loop is fed by a "Belief" and a highlighted "Hypothesis Injection" module, and it outputs a "Policy." In the top-center, a simple cycle of four ovals illustrates the core feedback loop: Action leads to Environment, which leads to Observation, which leads to Belief, and back to Action. On the top-right, a light yellow box titled "Belief Updating" shows how a "Belief" becomes a "Predicted Belief," which is corrected after conflicting with an "Observation." A key component in this correction phase is a highlighted "Conditional Reset" module, which feeds into a "Filter" to produce the new Belief. The bottom section contains two panels that provide visual examples. The panel on the bottom-left, titled "Conditional Reset," illustrates a scenario on a two-lane road where a truck occludes a pedestrian. It depicts a three-step process. Step 1 shows a belief about the pedestrian's current position, labeled as "s-hat with subscript p-comma-zero and superscript n," being projected forward. Step 2 shows a conflict, marked with a red X, where the predicted future position is in an occluded area. Step 3 shows the belief being reset to its original position, preserving the hypothesis of the hidden pedestrian. The panel on the bottom-right, titled "Hypothesis Injection," shows a car planning a trajectory to avoid a pedestrian. An inset box details the hypotheses being considered: "Surge," "Reverse," and "Freeze," each with a corresponding pedestrian icon and motion arrow. At the very bottom, a legend defines the graphical elements: A red-filled circle is "Belief of current pedestrian position." An orange-outlined circle is "Belief of future pedestrian position." Blue bars are "Belief of current pedestrian velocity." Teal bars are "Belief of future pedestrian velocity." A thick green arrow is "Planned future trajectory."}]{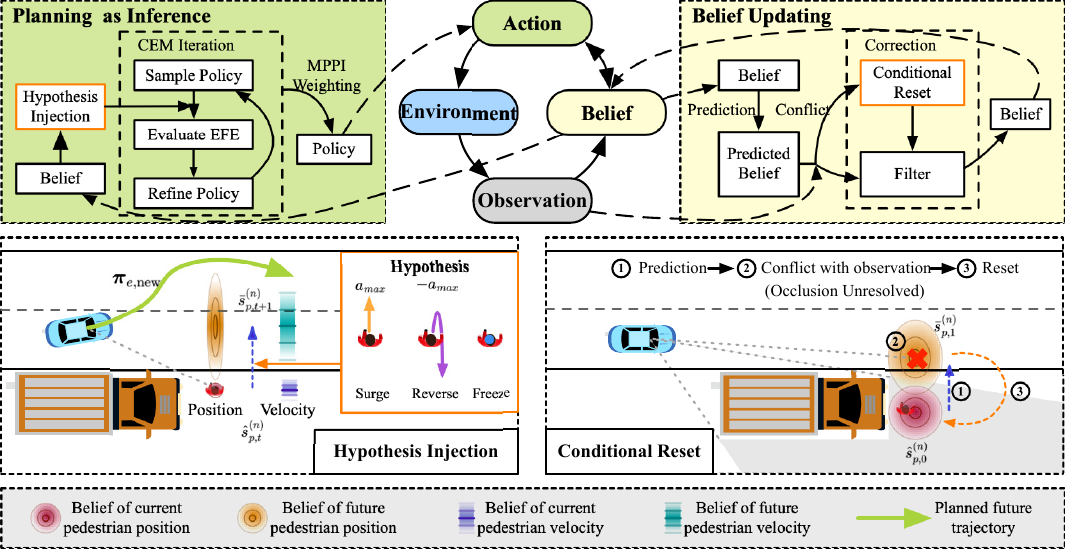}
  \caption{Overview of the Active Inference Interaction Framework. Our framework enhances the core perception-action loop (top center) with two key mechanisms to facilitate a explainable interaction. For robust \textit{Belief Updating} (top-right), \textbf{Conditional Reset} (bottom-left) prevents belief decay from occlusion. For proactive \textit{Planning} (top-left), \textbf{Hypothesis Injection} (bottom-right) challenges the agent with worst-case scenarios, yielding cautious and belief-driven behavior.}
  \Description[Overview of the Active Inference Interaction Framework.]{}
  \label{fig:framework}
\end{figure*}

Our framework operationalizes Active Inference through the continuous perception-action loop depicted in Figure~\ref{fig:framework}. This cycle enables the ego-vehicle to interact intelligently with potential pedestrians by constantly updating its internal \textit{Belief} from new \textit{Observations} and using that belief to select \textit{Actions} that shape future outcomes. 

The agent's belief $B_t$ over hidden states is approximated by a set of $N$ weighted particles $\{(w_t^{(n)}, \hat{s}_t^{(n)})_{n=1}^N\}$. These are propagated via an internal world model, comprising a transition model $P(s' \mid s, a)$ to predict state evolution and an observation model $P(o \mid s)$ to predict sensory outcomes. Each particle's state vector $\hat{s}_t^{(n)}$ and corresponding predicted observation vector $\hat{o}_t^{(n)}$ are defined as:
\begin{equation}
    \hat{s}_t^{(n)} = \left(\hat{s}_{e,t}^{(n)}, \hat{s}_{p,t}^{(n)}, \hat{s}_{p,0}^{(n)}, \hat{z}_{p}^{(n)}\right), \
    \hat{o}_t^{(n)} = \left(\hat{o}_{e,t}^{(n)}, \hat{o}_{p,t}^{(n)}, \hat{o}_{c,t}^{(n)}, \hat{o}_{r,t}^{(n)}\right)
\end{equation}
The state vector $\hat{s}_t^{(n)}$ contains the ego-vehicle state $\hat{s}_{e,t}^{(n)}$, the pedestrian's current kinematic state $\hat{s}_{p,t}^{(n)}$, and a binary variable for the pedestrian's existence, $\hat{z}_{p}^{(n)} \in \{0, 1\}$. Crucially, it also stores the particle's initial kinematic state hypothesis, $\hat{s}_{p,0}^{(n)}$, as a stable anchor for belief maintenance. The latent behavioral parameters $\theta_p$ (defined in Sec.~\ref{subsec:problem}) are not explicitly estimated; instead, their influence is captured through the stochastic transition model $P(s' \mid s, a)$, allowing each particle to represent a distinct behavioral hypothesis. The predicted observation vector includes the ego-vehicle's own state $\hat{o}_{e,t}^{(n)}$, pedestrian visibility $\hat{o}_{p,t}^{(n)}$, collision status $\hat{o}_{c,t}^{(n)}$, and a crucial occlusion resolution status $\hat{o}_{r,t}^{(n)}$. This status is true if the pedestrian is currently visible or if their initial hypothesized position is visible:
\begin{equation}
    \hat{o}_{p,t}^{(n)}\! =\! f_v\left(\hat{s}_{e,t}^{(n)}, \hat{s}_{p,t}^{(n)}, \mathcal{M}_{\mathrm{occ}}\right),
    \hat{o}_{r,t}^{(n)}\! =\! \hat{o}_{p,t}^{(n)}\! \lor\! f_v\left(\hat{s}_{e,t}^{(n)}, \hat{s}_{p,0}^{(n)}, \mathcal{M}_{\mathrm{occ}}\right) 
\end{equation}
where $f_v$ is a visibility checking function. 
Following the structure of Rao-Blackwellized particle filter (RBPF)~\cite{doucet2000rbpf}, the belief over the continuous kinematic state $\hat{s}_{p,t}^{(n)}$ conditioned on the discrete hypothesis (such as $\hat{z}_{p}^{(n)}$) is represented by a Gaussian distribution $\mathcal{N}(\boldsymbol{\mu}_{p, t}^{(n)}, \boldsymbol{\Sigma}_{p, t}^{(n)})$ managed by a dedicated Kalman Filter~\cite{kalman1960kf}.

This detailed belief representation serves as the foundation for the dual processes of the Active Inference loop. For \textit{Planning as Inference} (Section \ref{sec:planning}), the agent selects actions to minimize EFE. To foster \rev{a explainable} and safe human-robot interaction, this process is augmented with \textit{Hypothesis Injection}, emulating a defensive driver's \textit{what-if} reasoning about worst-case pedestrian behaviors. The resulting action influences the environment, yielding a new observation. It then drives \textit{Belief Updating} (Section \ref{sec:belief_updating}), which refines the agent's belief by minimizing VFE. Both the RBPF and its embedded Kalman Filter are practical mechanisms for this VFE minimization. This step incorporates a \textit{Conditional Reset} mechanism, mirroring a human driver's persistent vigilance by preventing the belief in a potential pedestrian from decaying simply due to non-observation. This continuous cycle allows the agent to navigate the interaction not as a simple avoidance task, but as a proactive, belief-aware partner that dynamically balances safety and efficiency.

\subsection{Belief Updating with Conditional Reset}
\label{sec:belief_updating}
The agent updates its belief from $B_{t-1}$ to $B_t$ using the action $a_{e, t-1}$ and the latest observation $o_t$. As illustrated in the \textit{Belief Updating} panel of Figure~\ref{fig:framework}, this process involves a prediction and a correction step. A conflict between the predicted belief and the observation can trigger a conditional reset mechanism. The observation $o_t = (o_{p, t}, \boldsymbol{x}_{p,t}, o_{c,t})$ is defined in Section \ref{sec:observation}.

\subsubsection{Prediction Step}
First, a prior belief $\bar{B}_t = \{(\bar{w}_t^{(n)}, \bar{s}_t^{(n)})\}_{n=1}^N$ is formed by propagating each particle from $B_{t-1}$ through the transition model $P(s' \mid s, a)$. The particle weights are carried over, such that $\bar{w}_t^{(n)} = w_{t-1}^{(n)}$. This step also yields the predicted observation $\bar{o}_t^{(n)}$ by observation model $P(o \mid s)$ for each particle, including the crucial occlusion resolution flag $\bar{o}_{r,t}^{(n)}$.

\subsubsection{Correction Step}
The correction phase updates the prior belief $\bar{B}_t$ to the posterior $B_t$. This process begins with a conditional state reset, followed by a unified state update step.

First, a conditional reset is applied to the kinematic state of each particle. Using an indicator function $\mathbb{I}(\cdot)$ that equals one if its condition is met and zero otherwise, the kinematic component $\bar{s}_{p,t}^{(n)}$ is updated to a modified prior state $\bar{s}_{p,t}'^{(n)}$ as follows:
\begin{equation}
    \bar{s}_{p,t}'^{(n)} = (1 - u) \cdot \bar{s}_{p,t}^{(n)} + u \cdot \hat{s}_{p,0}^{(n)}, \text{ where } u =  \mathbb{I}(o_{p,t}=0, \bar{o}_{r,t}^{(n)}=0)
\end{equation}
This operation resets the state to its initial hypothesis $\hat{s}_{p,0}^{(n)}$ only when the pedestrian is unobserved ($o_{p,t}=0$) and the particle's hypothesized occlusion zone is unresolved ($\bar{o}_{r,t}^{(n)}=0$). As visualized in the \textit{Conditional Reset} panel of Figure~\ref{fig:framework}, this mechanism is crucial for belief preservation under uncertainty. It addresses the conflict where a particle's prediction (\textcircled{1}) is inconsistent with the observation (\textcircled{2}) due to unresolved occlusion. Without this step, the weights of particles upholding the pedestrian's existence would be incorrectly penalized. The reset operation preempts this by reverting the particle's state to its initial hypothesis (\textcircled{3}), thus protecting these crucial particles from being erroneously discarded.

Crucially, a new predicted observation $\bar{o}_t'^{(n)}$ is generated from the modified prior state $\bar{s}_t'^{(n)} = \left(\bar{s}_{e,t}^{(n)}, \bar{s}_{p,t}'^{(n)}, \hat{s}_{p,0}^{(n)}, \bar{z}_{p}^{(n)}\right)$. This yields a corrected occlusion resolution flag, $\bar{o}_{r,t}'^{(n)}$. The posterior weights $w_t^{(n)}$ are calculated as follows~\cite{gordon1993pf, doucet2001pf}:
\begin{equation}
    w_t^{(n)} \propto \bar{w}_t^{(n)} \cdot P(o_t \mid \bar{s}_t'^{(n)})
\end{equation}
where the likelihood function is defined as:
\begin{equation}
    P(o_t \mid \bar{s}_t'^{(n)}) \propto P_{\mathrm{exist}}(o_{p,t} \mid \bar{z}_{p}^{(n)}, \bar{o}_{r,t}'^{(n)}) \cdot P_{\mathrm{measure}}(o_t \mid \bar{s}_t'^{(n)})
\end{equation}
The first term, $P_{\mathrm{exist}}$, evaluates the consistency between the particle's existence belief $\bar{z}_{p}^{(n)}$ and the observation flag $o_{p,t}$, conditioned on whether the occlusion is resolved ($\bar{o}_{r,t}'^{(n)}=1$). The second term, $P_{\mathrm{measure}}$, incorporates the measurement likelihoods only when the pedestrian is observed ($o_{p,t}=1$), which can be expressed compactly using the observation flag as an exponent:
\begin{equation}
    P_{\mathrm{measure}}(o_t \mid \bar{s}_t'^{(n)}) = \left[ P(\boldsymbol{x}_{p,t} \mid \bar{s}_{p,t}'^{(n)}) \cdot P(o_{c,t} \mid \bar{s}_{t}'^{(n)}) \right]^{o_{p,t}}
\end{equation}
Here, $P(\boldsymbol{x}_{p,t} \mid \bar{s}_{p,t}'^{(n)})$ is the kinematic likelihood from the Kalman filter, and $P(o_{c,t} \mid \bar{s}_{t}'^{(n)})$ is the likelihood of the collision status.

Finally, the posterior kinematic state $\hat{s}_{p,t}^{(n)}$ is determined. If the pedestrian is observed ($o_{p,t} =1$), the Kalman filter~\cite{kalman1960kf} performs a correction step on the modified prior state $\bar{s}_{p,t}'^{(n)}$ using the measurement $\boldsymbol{x}_{p,t}$. If not observed ($o_{p,t}=0$), the state simply carries over from the conditional reset step: $\hat{s}_{p,t}^{(n)} = \bar{s}_{p,t}'^{(n)}$.

\subsubsection{Resampling}
To mitigate particle degeneracy, we employ systematic resampling~\cite{arulampalam2002resample}. This step re-populates the particle set based on posterior weights, ensuring that computational effort is concentrated on the most plausible state hypotheses.

\subsection{Planning with Hypothesis Injection}
\label{sec:planning}

Planning is cast as an inference problem to find an optimal action sequence $\boldsymbol{\pi}_{e}$ that minimizes the EFE over a finite horizon $T$. The overall workflow, illustrated in the \textit{Planning as Inference} panel of Figure~\ref{fig:framework}, integrates a crucial defensive driving mechanism, \textit{Hypothesis Injection}, into an iterative MPPI-CEM planner. This allows the agent to reason not only about the most likely future, but also about low-probability, high-impact \textit{what-if} scenarios.

\subsubsection{Hypothesis Injection for Counterfactual Reasoning}

To implement the cognitive process of \textit{what-if} reasoning, our planner relies not only on the current belief $B_t$. Instead, it uses \textit{Hypothesis Injection} to generate a temporary, augmented belief $B'_t$ used exclusively for planning rollouts. As depicted in the \textit{Hypothesis Injection} panel of Figure~\ref{fig:framework}, this mechanism imagines a set of counterfactual pedestrian behaviors, such as an abrupt \textit{Surge}, \textit{Reverse}, or \textit{Freeze}.

Operationally, these abnormal maneuver modes are assigned to a small fraction ($\rho_H$) of the particles. The remaining majority of particles continue to follow the normal predictive model. This creates a diverse and challenging set of imagined scenarios, forcing the planner to evaluate its candidate trajectories against worst-case possibilities. Crucially, this entire process is confined to the planner's rollouts and does not corrupt the agent's persistent, factual belief $B_t$, which is updated separately. The resulting temporary planning belief is $B'_t = \{(w_t^{(n)}, (\hat{s}_t^{(n)}, m^{(n)}))\}_{n=1}^N$, where $m^{(n)}$ denotes the assigned maneuver mode for each particle.

\paragraph{Planning with MPPI-CEM}

We employ a sampling-based model predictive control scheme that seeks an optimal policy by minimizing the EFE. At each time step, our planner samples a set of $M$ candidate policies $\{\pi_e^{(m)}\}_{m=1}^M$. \rev{Each policy $\pi_e^{(m)} = (a_{e,t}^{(m)}, \dots, a_{e,t+T-1}^{(m)})$ is a sequence of actions over a planning horizon of $T$ steps. The total EFE for a given policy, $G(\pi_e^{(m)})$, is the discounted sum of single-step EFEs.}
\rev{To practically compute the single-step EFE $G_k(\pi_e^{(m)})$ for $k \in \{t+1, \dots, t+T\}$, we apply the simplification for deterministic observation models (derived in Appendix~\ref{app:A}) and approximate it via a rollout of each particle in the current belief $B_t$.}
This rollout process  generates a trajectory of predicted states  $\bar{s}_{k}^{(n)}$  and observations $\bar{o}_{k}^{(n)}$ for each particle, from which the EFE is approximated:
\begin{equation}
    G_k(\pi_e^{(m)})\! \approx\! -\!\underbrace{ \sum_{n=1}^N w_t^{(n)}\! \left( \log P_d(\bar{o}_{e,k}^{(n)},\bar{o}_{c,k}^{(n)})\right) }_{\text{Pragmatic Value}} \\
     -\underbrace{H\!\left[ \sum_{n=1}^N w_t^{(n)} \delta(\cdot - \bar{o}_{p,k}^{(n)}) \right] }_{\text{Epistemic Value}}
\end{equation}
where $P_d$ represents the preference distribution over desired outcomes. It encodes the preference for desired driving behavior (related to $\bar{o}_{e,k}$) and for avoiding collisions (related to $\bar{o}_{c,k}$). The epistemic value encourages information gathering to resolve uncertainty about the pedestrian's visibility (observed via $\bar{o}_{p,k}$). This approximation to visibility entropy is justified as our deterministic observation model collapses kinematic uncertainty upon detection, making visibility itself the primary and most computationally tractable epistemic objective.

\rev{Our planner employs a hybrid MPPI-CEM approach. It first uses several CEM (Cross Entropy Method~\cite{rubinstein1999cem_paper, rubinstein2004cem_book}) iterations to efficiently refine the policy sampling distribution towards low-cost regions. The optimal policy $\boldsymbol{\pi}_{e}^*$ is then computed via an MPPI~\cite{williams2016mppi} weighted average, where weights are proportional to the exponentiated negative cost $\exp(-G(\pi_e^{(m)})/\lambda)$. The first action of the policy, $a_{e,t}^*$, is then executed, and the process repeats at the next time step.}

\section{Empirical Results}

\subsection{Experimental Setup}

To evaluate the performance of our proposed framework, we designed a challenging Gym~\cite{brockman2016gym} simulation environment using the Gymnasium~\cite{towers2024gymnasium} API, centered on the occluded pedestrian scenario defined in Section \ref{subsec:problem}.

\subsubsection{Simulation Environment}
\rev{While the ego-vehicle starts with a fixed initial state, the pedestrian's behavior is highly stochastic. To cover a wide range of corner cases, we implement five distinct and challenging pedestrian behavioral modes:
\begin{enumerate}[leftmargin=14pt] 
    \item \textbf{Hesitant Pedestrian:} The pedestrian enters the road but exhibits indecisiveness, randomly switching between moving, stopping, and slightly reversing.
    \item \textbf{Deceptive Accelerating Pedestrian:} The pedestrian initially moves slowly but executes a sudden, aggressive acceleration when the vehicle is in close proximity.
    \item \textbf{Turning-Back Pedestrian:} The pedestrian starts crossing but suddenly turns back and returns to the sidewalk mid-crossing.
    \item \textbf{Sudden Stop Pedestrian:} The pedestrian initiates a crossing but abruptly halts in the middle of the road.
    \item \textbf{Sudden Appearance Pedestrian:} The pedestrian rushes out from behind the occluder with high initial velocity and acceleration when the vehicle is already in close proximity.
\end{enumerate}}

\rev{To ensure behavioral diversity within each mode, we randomize a comprehensive set of parameters for every episode, including initial positions, target velocities, acceleration capabilities, hesitation durations, and turning points, among others. Detailed parameter distributions and kinematic settings are provided in Appendix~\ref{app:B}.}

\subsubsection{Evaluation Metrics}
We assess performance using five key metrics, with results averaged over 600 runs and reported as mean $\pm$ standard error (SE).
\textbf{Collision Rate (CR)} is the percentage of runs ending in a collision, serving as the primary safety metric.
\textbf{Pass Rate (PR)} is the percentage of trials where the vehicle successfully avoids a collision.
\textbf{Pass Time (PT)} measures efficiency as the average time to complete successful trials.
To assess the safety margin, we use \textbf{Minimum Distance (MD)}, the average minimum Euclidean distance to the pedestrian in successful runs.
Finally, \textbf{Minimum TTC (TTC)}~\cite{hayward1972ttc} quantifies temporal risk as the average minimum time until a potential collision at current velocities, also measured on successful trials.

\subsection{Ablation Studies}

\begin{table*}
  \caption{Influence of Initial Presence Belief ( $B_0(z_p)$) and Hypothesis Injection Ratio ($\rho_{H}$) (Metrics are averaged across 5 scenarios).} 
  \label{tab:ablation_study}
  \begin{tabular}{c c c c c c c c}
  \toprule
   $\rho_H$ & $B_0(z_p)$ & \textbf{Pass Rate (PR)} $\uparrow$  & \textbf{Collision Rate (CR)} $\downarrow$ & \textbf{Pass Time (PT)} $\downarrow$ & \textbf{Min. Distance (MD)} $\uparrow$ & \textbf{Min. TTC (TTC)} $\uparrow$ \\
  \midrule
  0.0 & 0.8 & $0.905 \pm 0.012$ & $0.095 \pm 0.012$ & $\textbf{7.974} \pm \textbf{0.026}$ & $3.679 \pm 0.049$ & $\textbf{1.739} \pm \textbf{0.041}$ \\
  \addlinespace
  0.1 & 0.8 & $0.923 \pm 0.011$ & $0.077 \pm 0.011$ & $8.022 \pm 0.029$ & $3.857 \pm 0.049$ & $1.734 \pm 0.036$ \\
  \addlinespace
  0.3 & 0.8 & $0.947 \pm 0.010$ & $0.053 \pm 0.010$ & $8.653 \pm 0.052$ & $\textbf{3.890} \pm \textbf{0.041}$ & $1.540 \pm 0.030$ \\
  \addlinespace
  0.8 & 0.8 & $\textbf{0.977} \pm \textbf{0.007}$ & $\textbf{0.023} \pm \textbf{0.007}$ & $9.148 \pm 0.064$ & $3.367 \pm 0.024$ & $0.963 \pm 0.016$ \\
  \midrule
  0.3 & 0.0 & $0.250 \pm 0.018$ & $0.750 \pm 0.018$ & $\textbf{6.448} \pm \textbf{0.007}$ & $3.770 \pm 0.084$ & $1.220 \pm 0.045$ \\
  \addlinespace
  0.3 & 0.2 & $0.707 \pm 0.019$ & $0.293 \pm 0.019$ & $7.141 \pm 0.029$ & $3.572 \pm 0.042$ & $1.013 \pm 0.025$ \\
  \addlinespace
  0.3 & 0.4 & $0.880 \pm 0.013$ & $0.120 \pm 0.013$ & $7.790 \pm 0.041$ & $3.695 \pm 0.038$ & $1.158 \pm 0.023$ \\
  \addlinespace
  0.3 & 0.8 & $\textbf{0.947} \pm \textbf{0.010}$ & $\textbf{0.053} \pm \textbf{0.010}$ & $8.653 \pm 0.052$ & $\textbf{3.890} \pm \textbf{0.041}$ & $\textbf{1.540} \pm \textbf{0.030}$ \\
  \bottomrule
  \end{tabular}
\end{table*}

\subsubsection{Conditional Reset.}
To highlight the importance of maintaining caution under uncertainty, we analyze an interactive occlusion scenario with and without our \textit{Conditional Reset} mechanism. This mechanism is designed to prevent the agent's belief in a potential but unobserved pedestrian from decaying over time.

\begin{figure}[ht]
  \centering
  \includegraphics[width=\linewidth, alt={A figure containing two line plots, labeled (a) and (b), comparing two conditions: "Without Conditional Reset" shown as a blue dashed line with circle markers, and "With Conditional Reset" shown as an orange dashed line with triangle markers.
    Plot (a) shows "B(zp) versus Time (s)". The y-axis, B(zp), ranges from 0.00 to 1.00. The x-axis, Time, ranges from 0.0 to 10.0 seconds. The blue line starts near 0.4, decreases to 0 at 3 seconds, and then sharply jumps to 1.0 at approximately 4.2 seconds, where a vertical line indicates a "Collision". The orange line starts near 0.4 and remains constant until it also jumps to 1.0 at 4.2 seconds. A second vertical line near 9.5 seconds indicates "Goal Reached" for the orange trajectory.
    Plot (b) shows "Speed (m/s) versus Time (s)". The y-axis, Speed, ranges from 0.0 to 12.5 m/s. The blue line starts around 10 m/s, dips slightly, then rises to a peak above 12.5 m/s at the 4.2-second "Collision" line, where the data ends. The orange line starts around 10.5 m/s and steadily decreases to a minimum of about 2.5 m/s at 4.2 seconds. After this point, the speed recovers and increases toward the "Goal Reached" event.}]{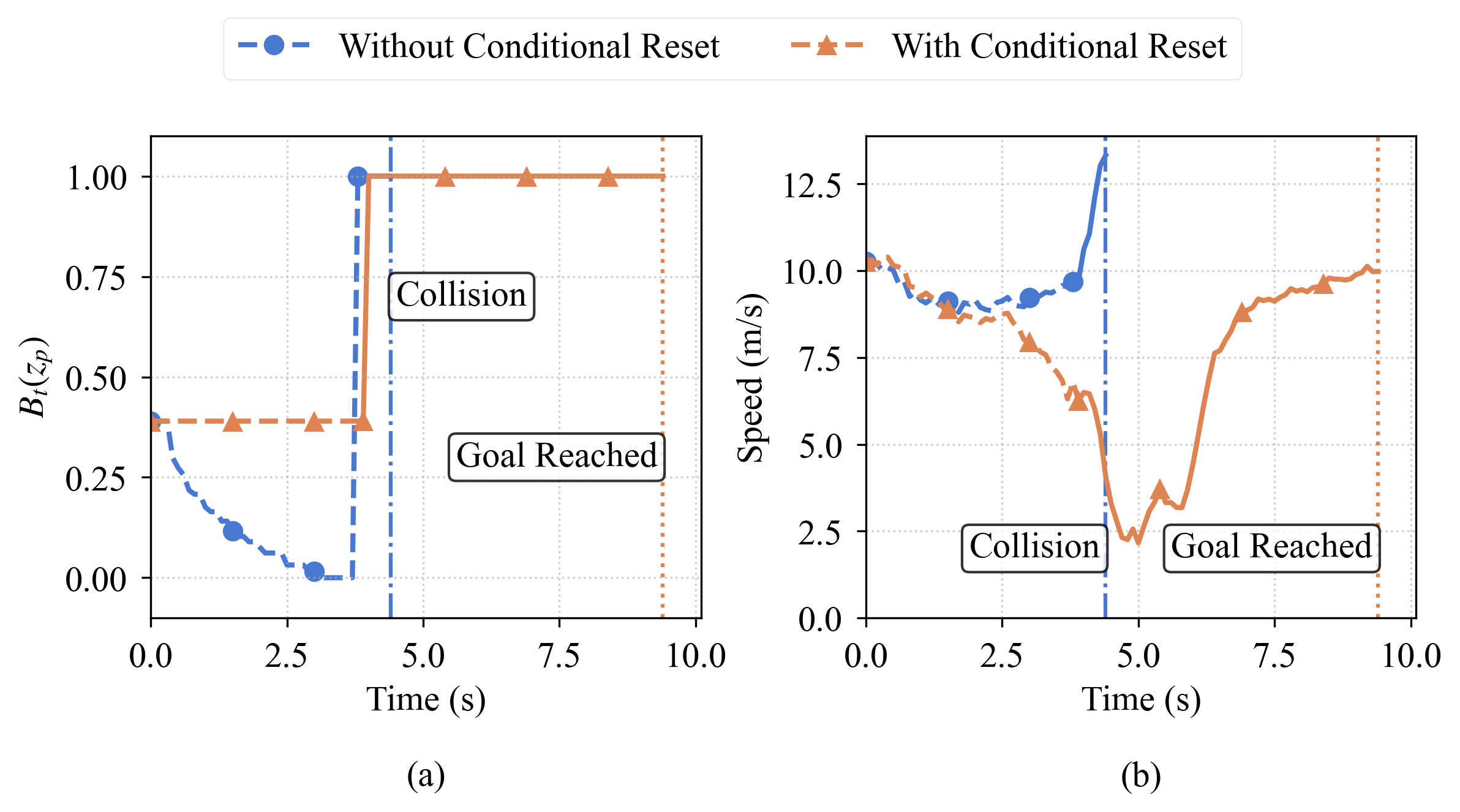}
  \Description{}
  \caption{Impact of Conditional Reset in an Occlusion Scenario.
    (a) Belief in the pedestrian's existence, $B_t(z_p)=\sum_n w_t^{(n)} \hat{z}_p^{(n)}$, and (b) the agent's speed over time.
    The agent without Conditional Reset (blue) allows its belief to decay to zero, leading to risky acceleration and a collision.
    In contrast, the agent with Conditional Reset (orange) maintains its belief, decelerates cautiously, and safely navigates past the revealed pedestrian to reach its goal.}
  \label{fig:particles_reset}
\end{figure}

As shown in Figure \ref{fig:particles_reset}, the outcomes differ dramatically. The agent without Conditional Reset (blue curves) allows its belief in the pedestrian to decay to zero while the view is occluded (Figure \ref{fig:particles_reset}a). This erroneous belief causes it to accelerate into the uncertain area (Figure \ref{fig:particles_reset}b). When the pedestrian is finally observed at $t \approx 4.2~\mathrm{s}$, the agent is moving too fast to react, resulting in a collision. In contrast, the agent equipped with Conditional Reset (orange curves) maintains a persistent, non-zero belief about the potential hazard. This sustained caution compels it to decelerate as it approaches the occlusion, reaching a safe speed before the pedestrian is revealed. This cautious behavior allows it to easily avoid a collision and proceed to its goal, demonstrating that the mechanism is critical for safe interaction in uncertain environments.

\subsubsection{Hypothesis Injection.}

\begin{figure}[ht]
  \centering
  \includegraphics[width=\linewidth, alt={A figure containing two plots, labeled (a) and (b), sharing a common legend at the top. The legend shows four conditions: rho_H=0.0 (blue line with circle marker), rho_H=0.1 (light blue line with triangle marker), rho_H=0.3 (green line with square marker), and rho_H=0.8 (tan line with plus marker). Plot (a) is a top-down view of a driving scenario. The axes are "X Position (m)" from -5 to 15 and "Y Position (m)" from -10 to 5. An orange truck is depicted near the center. Four colored lines show vehicle trajectories starting from the left. The blue line proceeds almost straight and ends at a red 'X' marker, indicating a collision with a red pedestrian icon emerging from behind the truck. The other lines (light blue, green, tan) swerve progressively more to the left, creating more space around the pedestrian. Plot (b) is a line graph of "Speed (m/s)" versus "Time (s)". The y-axis ranges from 0 to 12.5, and the x-axis from 0 to 8. All lines start around 10 m/s and show a dip in speed. The blue line dips shallowly before rising and abruptly ending at a vertical dash-dotted line labeled "Collision" around 4.8 seconds. The other three lines show a much deeper dip in speed, reaching a minimum of around 2.5 m/s, before recovering.}]{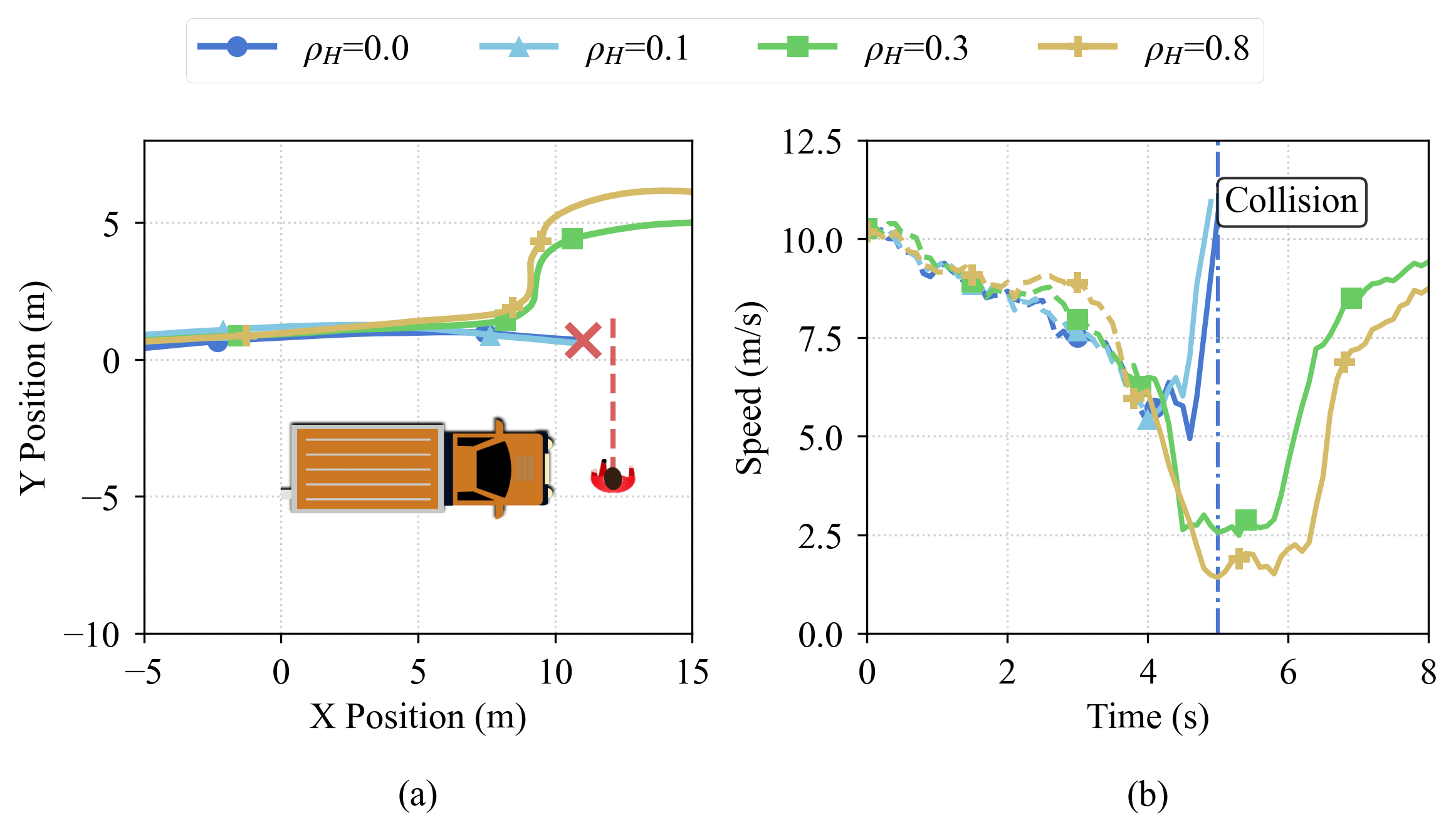}
  \caption{Effect of Hypothesis Injection Ratio ($\rho_H$) on Behavior. 
  (a) Agent's trajectory and (b) speed profile for different $\rho_H$. A low ratio ($\rho_H=0.0$) leads to insufficient deceleration and a collision. As $\rho_H$ increases, the agent plans a more cautious trajectory with a wider berth (a) by decelerating more significantly (b), thus successfully avoiding the pedestrian.}
  \Description{}
  \label{fig:manuver_speed}
\end{figure}

We evaluate the impact of the Hypothesis Injection Ratio ($\rho_H$), which controls the diversity of pedestrian motion hypotheses the agent considers during planning. As illustrated qualitatively in Figure \ref{fig:manuver_speed} and confirmed quantitatively in Table~\ref{tab:ablation_study}, a higher $\rho_H$ value induces more cautious and safer behavior when the agent interacts with a partially occluded pedestrian.

Figure \ref{fig:manuver_speed} shows that an agent with no hypothesis injection ($\rho_H=0.0$, blue) plans only for the most likely outcome, becoming overconfident. This leads to insufficient deceleration and a subsequent collision (Figure \ref{fig:manuver_speed}b). In contrast, increasing $\rho_H$ forces the planner to account for more diverse, worst-case pedestrian movements. This proactive consideration results in more cautious behavior: the agent creates a wider safety margin by swerving further away (Figure \ref{fig:manuver_speed}a) and decelerates more aggressively to increase its reaction time.

This qualitative analysis is directly supported by the quantitative metrics in the top rows of Table~\ref{tab:ablation_study}. With the initial presence belief $B_0(z_p)$ fixed at $0.8$, increasing the Hypothesis Injection Ratio $\rho_H$ induces progressively more conservative driving behavior, significantly enhancing safety. Specifically, the \textbf{CR} decreases from $9.5\%$ at $\rho_H=0.0$ to just $2.3\%$ at $\rho_H=0.8$. This mechanism proves particularly critical in scenarios characterized by high dynamic uncertainty, where pedestrian motion states change abruptly. For instance, in the \textit{Turning-Back Pedestrian} scenario, the CR drops from $21.7\%$ ($\rho_H=0.0$) to $7.5\%$ ($\rho_H=0.3$). Similarly, for the \textit{Deceptive Accelerating Pedestrian}, the CR is drastically reduced from $19.2\%$ with $\rho_H=0.0$ to just $1.7\%$ with $\rho_H=0.8$. These results demonstrate that hypothesis injection is a highly effective mechanism for enhancing interaction safety by preparing the agent for worst-case evolutions of the environment. More details are provided in Appendix~\ref{app:C}.

\begin{figure}[ht]
  \centering
  \includegraphics[width=\linewidth, alt={A figure with two plots, (a) and (b), showing the effect of different B0(zp) values: 0.0 (blue), 0.2 (light blue), 0.4 (green), and 1.0 (tan). Plot (a) shows top-down trajectories. An orange truck occludes the view of a pedestrian. The blue line (B0=0.0) drives straight and collides with the pedestrian, marked by a red 'X'. The light blue, green, and tan lines swerve progressively further to the left to avoid the pedestrian, with the tan line (B0=1.0) creating the widest berth. Plot (b) shows speed versus time. The blue line remains at a constant high speed until the collision event, marked by a vertical line. The light blue line shows erratic speed before a sharp spike at the collision. The green and tan lines show smooth, early deceleration to a low speed around the time of potential interaction, followed by re-acceleration. The tan line shows the earliest and deepest deceleration.}]{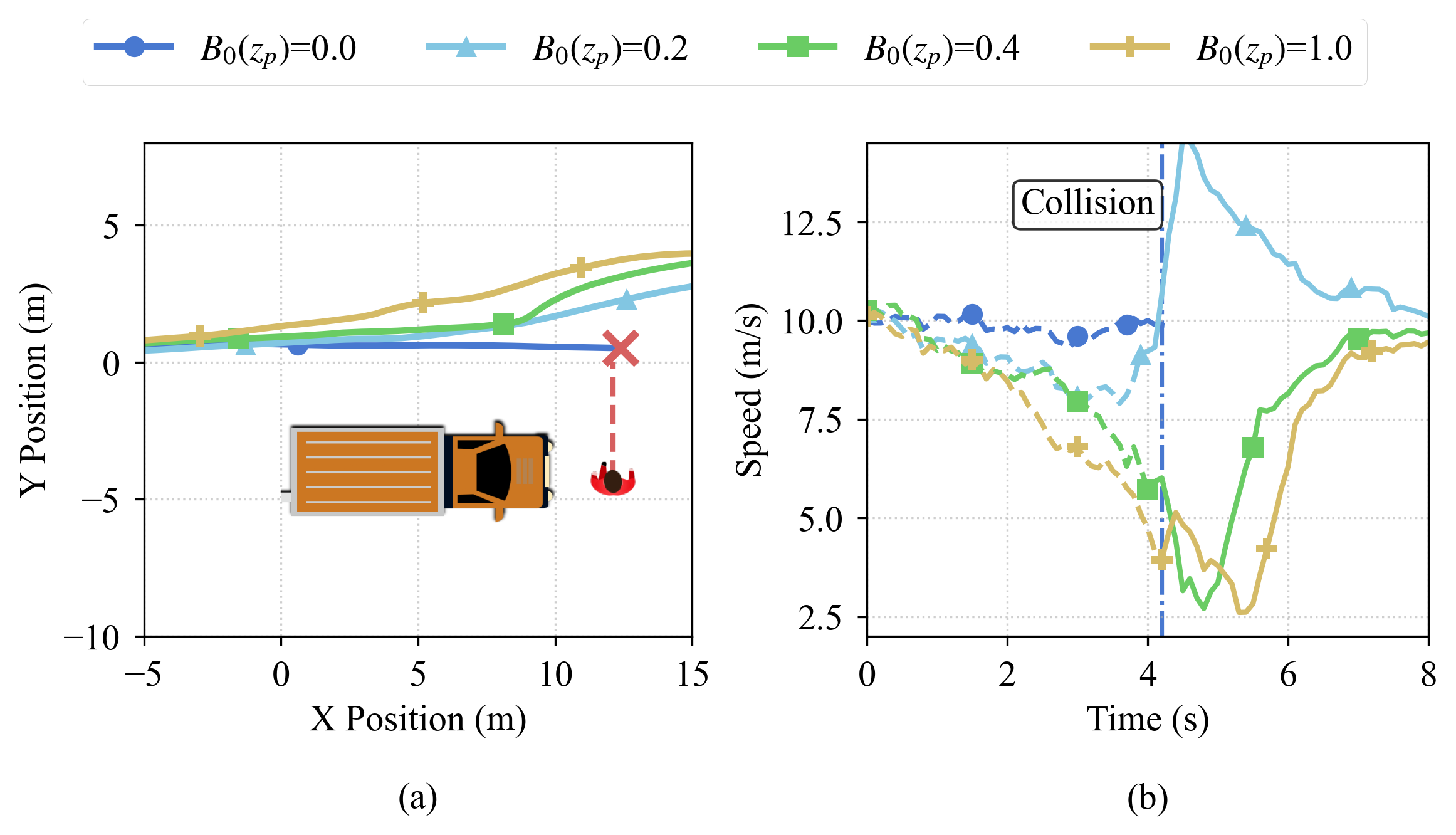}
  \caption{Impact of Initial Presence Belief ($B_0(z_p)$) on Behavior.
  (a) Agent's trajectory and (b) speed profile for different initial beliefs. A zero belief ($B_0(z_p)=0.0$) leads to no evasive action and a collision. As the belief increases, the agent decelerates earlier and more significantly, ensuring a safe passage.}
  \Description{}
  \label{fig:belief_speed}
\end{figure}

\subsubsection{Initial Presence Belief.}

We analyze the influence of the agent's prior conviction about the pedestrian's existence, the \textit{Initial Presence Belief} $B_0(z_p)$. It directly governs the agent's baseline level of caution in uncertain situations, effectively acting as a mechanism for risk modulation.

As illustrated in Figure \ref{fig:belief_speed}, the agent's behavior is highly sensitive to its initial belief. When the belief is zero ($B_0(z_p)=0.0$), the agent acts with complete overconfidence, maintaining high speed and making no attempt to slow down, leading to a preventable collision. As the initial belief increases, the agent exhibits progressively more cautious behavior. This is most clearly demonstrated in the speed profiles (Figure \ref{fig:belief_speed}b), where a higher $B_0(z_p)$ leads to both earlier and more significant deceleration in anticipation of the potential hazard. This speed reduction is the primary mechanism for ensuring safety. The wider lateral berth observed in trajectories with higher belief (Figure \ref{fig:belief_speed}a) is a natural consequence of this early cautious planning.

The bottom rows of Table~\ref{tab:ablation_study} illustrate the impact of the Initial Presence Belief $B_0(z_p)$ while the Hypothesis Injection Ratio is fixed at $\rho_H=0.3$. As $B_0(z_p)$ increases, the model exhibits progressively more conservative driving behavior, validating the effectiveness of the proposed method. Consequently, the overall \textbf{CR} drops significantly from $75.0\%$ at $B_0(z_p)=0.0$ to $5.3\%$ at $B_0(z_p)=0.8$. Scenario-specific breakdowns reveal that higher initial beliefs significantly enhance performance across all cases, particularly in scenarios characterized by high suddenness. In the \textit{Sudden Appearance Pedestrian} and \textit{Sudden Stop Pedestrian} scenarios, the absence of an initial belief leads to catastrophic failure, with CR of $100\%$ and $95.8\%$, respectively. Increasing $B_0(z_p)$ to $0.8$ mitigates these risks substantially, reducing CR to $0\%$ and $0.8\%$. In contrast, for the \textit{Hesitant Pedestrian} scenario, which involves lower suddenness and dynamic uncertainty, a slight increase in belief is sufficient; adjusting $B_0(z_p)$ from $0.0$ to $0.2$ causes the CR to plummet from $39.2\%$ to $3.3\%$. More details are provided in Appendix~\ref{app:C}.

\begin{table}
  \caption{\rev{Average Wall-clock Time (ms) per Planning Step for Varying Numbers of Particles ($N$) and Candidate Policies ($M$).}}
  \label{tab:computation_cost}
  \setlength{\tabcolsep}{3pt} 
  \begin{tabular}{c c c c c}
  \toprule
   \diagbox{$\mathbf{M}$}{$\mathbf{N}$} & $\mathbf{100}$ & $\mathbf{200}$  & $\mathbf{300}$ & $\mathbf{500}$ \\
   \midrule
   $\mathbf{100}$ & $17.69 \pm 0.17$ & $20.77 \pm 0.22$ & $23.64 \pm 0.18$ & $27.98 \pm 0.17$ \\
   \addlinespace
   $\mathbf{200}$ & $21.08 \pm 0.19$ & $24.49 \pm 0.13$ & $29.58 \pm 0.19$ & $52.15 \pm 0.10$ \\
   \addlinespace
   $\mathbf{400}$ & $24.35 \pm 0.19$ & $42.43 \pm 0.09$ & $58.63 \pm 0.11$ & $88.95 \pm 0.13$ \\
  \bottomrule
  \end{tabular}
\end{table}

\rev{\subsection{Computational Efficiency}
To assess the real-time feasibility of our framework, we quantified its computational latency. The framework is implemented using JAX~\cite{jax2018github} to leverage the parallel computing capabilities on an NVIDIA RTX 3090 GPU. We evaluated the average wall-clock time per planning step by varying two critical hyperparameters: the number of candidate policies sampled by the planner ($M$) and the number of particles used for belief representation ($N$). Throughout these tests, the number of CEM iterations was fixed at 8, and the planning horizon was set to 40 steps. To ensure statistical reliability, the reported times are derived from 10 independent repetitions, with each repetition averaging the runtime of 1,000 execution loops.}

\rev{The results are summarized in Table~\ref{tab:computation_cost}. For the configuration used in our main experiments ($M=100, N=100$), the average computation time is $17.69$~ms, leaving a substantial margin for other system processes. Even under the most demanding configuration with $N=500$ particles and $M=400$ candidate policies, the latency increases to only $88.95$~ms. This performance remains well within the $100$~ms threshold required for 10~Hz real-time control loops. These results demonstrate that our proposed method is computationally efficient and scalable, capable of handling complex interaction scenarios or broader policy searches.}

To evaluate the efficacy of our proposed belief-aware planning framework, we conduct a comparative analysis against three distinct baseline methods in the same challenging occluded crossing scenario. 
The results, summarized in Table~\ref{tab:baseline} and Figure~\ref{fig:baseline_comparison}, reveal that our method achieves a superior balance of safety and efficiency.

\subsection{Compared with Baseline methods}

\subsubsection{Baseline Methods}
Representative approaches include:

\subparagraph{\textbf{Reactive}}
This agent uses an MPPI planner that only reacts to visible threats. It lacks any mechanism to anticipate or reason about the occluded pedestrian before it becomes visible.

\subparagraph{\textbf{Rule-based}}
This baseline adds a fixed, hand-crafted safety rule to the Reactive planner. It is forced to decelerate to a constant low speed when approaching the occlusion, representing a common heuristic-based safety approach.

\subparagraph{\textbf{PPO-LSTM}}
This is a model-free reinforcement learning agent using PPO~\cite{schulman2017ppo} with an LSTM~\cite{hochreiter1997lstm, gers2000lstm} to process observation history. It must learn a policy from trial-and-error experience without an explicit world model or belief management.

\begin{figure}[ht]
  \centering
  \includegraphics[width=\linewidth, alt={A 2x2 grid of plots, each showing a top-down driving scene. The top-left, 'Reactive', shows two trajectories colliding with a pedestrian. The top-right, 'Rule-based', shows two slow, conservative trajectories. The bottom-left, 'PPO-LSTM', shows two nearly identical trajectories, one of which results in a collision. The bottom-right, 'Ours', shows two distinct, adaptive trajectories: a sharp blue curve for a late-appearing pedestrian and a smoother, more efficient yellow curve for an early-appearing one.}]{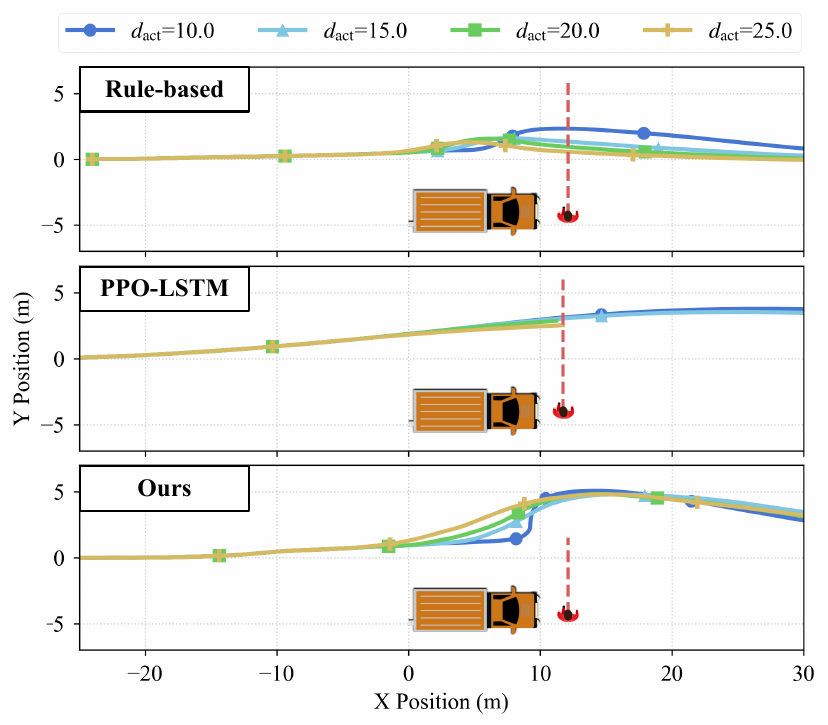}
  \caption{Qualitative Comparison of Driving Strategies. 
  The trajectories show how each method responds to the pedestrian appearing at different times.
  Baselines execute fixed, non-adaptive strategies. In contrast, our method (bottom right) dynamically adapts its maneuver to the perceived risk, demonstrating intelligent behavior.}
  \Description{}
  \label{fig:baseline_comparison}
\end{figure}

\begin{table*}
  \caption{\rev{Quantitative Comparison with Baseline Methods (Metrics are averaged across 5 scenarios).}}
  \label{tab:baseline}
  \begin{tabular}{l c c c c c c }
  \toprule
   \textbf{Method} & \textbf{Pass Rate (PR)} $\uparrow$  & \textbf{Collision Rate (CR)} $\downarrow$ & \textbf{Pass Time (PT)} $\downarrow$ & \textbf{Min. Distance (MD)} $\uparrow$ & \textbf{Min. TTC (TTC)} $\uparrow$ \\
  \midrule
  Reactive & $0.178 \pm 0.016$ & $0.822 \pm 0.016$ & $ 6.508 \pm 0.042$ & $3.050 \pm 0.083$ & $0.835 \pm 0.071$ \\
  \addlinespace
  Rule-based & $0.587 \pm 0.020$ & $0.413 \pm 0.020$ & $ 7.755 \pm 0.036$ & $2.971 \pm 0.043$ & $0.650 \pm 0.032$ \\
  \addlinespace
  PPO-LSTM & $0.725 \pm 0.018$ & $0.275 \pm 0.018$ & $\textbf{4.188} \pm \textbf{0.002}$ & $\textbf{4.080} \pm \textbf{0.038}$ & $\textbf{9.663} \pm \textbf{0.086}$ \\
  \midrule
  Ours & $\textbf{0.947} \pm \textbf{0.010}$ & $\textbf{0.053} \pm \textbf{0.010}$ & $8.653 \pm 0.052$ & $3.890 \pm 0.041$ & $1.540 \pm 0.030$ \\
  \bottomrule
  \end{tabular}
\end{table*}

\subsubsection{Qualitative Behavioral Analysis}
\label{sec:baseline_qualitative}
Figure~\ref{fig:baseline_comparison} reveals the differences in strategy. The baselines exhibit rigid, non-adaptive behaviors. The PPO-LSTM agent, for instance, converges to a single generic policy, executing nearly the same trajectory regardless of when the pedestrian appears. This explains its brittleness and high failure rate when its generalized policy does not fit a specific high-risk encounter. In contrast, our method shows both proactive and adaptive behavior. It proactively maneuvers to improve its field of view, and further tailors its response to the situation. When the pedestrian appears late (blue curve), signaling high risk, our agent executes a sharp, decisive evasive maneuver. When the pedestrian appears early (yellow curve), it correctly identifies a lower risk and plots a much smoother, more efficient path. This ability to dynamically modulate its defensive posture based on the evolving uncertainty is the key to its superior performance.

\subsubsection{Quantitative Analysis}
Table~\ref{tab:baseline} presents the quantitative comparison with baseline methods. Our method demonstrates superior safety performance compared to all baselines, achieving a \textbf{CR} of only $5.3\%$, which is significantly lower than that of the Rule-based method ($41.3\%$) and PPO-LSTM ($27.5\%$). Scenario-specific experiments reveal that the Rule-based method performs relatively well in the \textit{Sudden Stop Pedestrian} and \textit{Hesitant Pedestrian} scenarios, with CRs of $10.8\%$ and $15.0\%$, respectively. However, it incurs a high CR of $86.7\%$ in the \textit{Sudden Appearance Pedestrian} scenario, indicating that relying solely on a simple deceleration rule is insufficient for handling all situations. Conversely, PPO-LSTM excels only in the \textit{Sudden Appearance Pedestrian} scenario ($0\%$ CR). This observation aligns with the qualitative analysis in Section \ref{sec:baseline_qualitative}, suggesting that the model has primarily learned a pattern of bypassing obstacles for rapid traversal but fails to handle complex behavioral dynamics. Detailed results are provided in Appendix~\ref{app:D}.

\section{Discussion}

\subsection{Pathway to Deployment}
A potential deployment of our framework within a modular autonomous driving system would involve an upstream perception module that processes sensor data to detect objects like the occluding bus. Crucially, this is augmented by a sophisticated scene understanding model, such as a Vision-Language Model (VLM)~\cite{pan2024vlp, zhou2024vlm_survey}, to provide high-level semantic context. This VLM's role is to analyze the scene and infer the probability of latent hazards—for instance, assigning a high probability to an occluded pedestrian scenario upon seeing a bus stopped at a crosswalk. This output directly serves as the Initial Presence Belief ($B_0(z_p)$) for our planner. Our belief-aware planning module then uses this structured, probabilistic world representation and computes a kinematically feasible trajectory that optimally balances safety and efficiency.
Finally, this trajectory is passed to a downstream low-level controller for execution via precise steering and throttle commands. A key strength of this architecture is its two intuitive parameters for calibrating risk sensitivity. The VLM-generated $B_0(z_p)$ allows the vehicle's caution to dynamically adapt to real-time perceptual context. Concurrently, the Hypothesis Injection Ratio ($\rho_H$) acts as a configurable parameter for system designers to set the planner's intrinsic defensiveness, tuning how proactively it plans for potential pedestrian movements. This dual-parameter system enables manufacturers to define distinct driving profiles to align the vehicle's interactive behavior with specific safety requirements and user preferences.

\subsection{Limitations and Future Work}
\label{sec:limitations}

\rev{While our framework shows promise in occluded scenarios, several limitations remain. First, our evaluation uses a limited set of baselines. Future work will benchmark against a broader range of state-of-the-art interaction-aware methods for a more comprehensive assessment.
Second, unlike formal reachability analysis, our probabilistic framework does not offer absolute theoretical safety guarantees. However, by incorporating pedestrian kinematic limits into the generative model, we ensure a high degree of engineering safety suitable for practical use.
Third, our experiments were conducted in idealized simulations. Nevertheless, the integration of the Kalman Filter provides resilience to uncertainty. Preliminary tests indicate robustness against observation noise, suggesting feasibility for real-world transfer.
Looking forward, we aim to enhance interactive capabilities by incorporating game-theoretic concepts into the active inference loop. Furthermore, we plan to conduct user studies to evaluate social acceptability and comfort. Finally, we intend to bridge the sim-to-real gap by deploying the framework on a physical vehicle to validate performance in real traffic.}

\section{Conclusion}
\label{sec:conclusion}

In this paper, we addressed the critical challenge of safe and \rev{interpretable} vehicle-pedestrian interaction in uncertain occlusion scenarios. We proposed a novel framework grounded in active inference that enables an agent to perform human-like, belief-driven proactive decision-making. Our approach integrates a robust belief updating module with \textit{Conditional Reset} to maintain vigilance, and a proactive planning process with \textit{Hypothesis Injection} for counterfactual reasoning.
Our extensive simulations demonstrated that the framework significantly outperforms reactive, rule-based, and model-free RL baselines in both safety and efficiency. Crucially, the resulting behavior is not only safer but also more \rev{explainable}, as the agent's actions directly reflect its internal beliefs about latent risks. This work marks a significant step towards autonomous systems that can navigate complex human environments in a trustworthy manner, fostering safer human-robot interaction.

\section*{Data-Availability Statement}
The Python implementation of the Active Inference Interaction Framework, the simulation environment featuring all pedestrian behavioral modes, and all baseline methods are available in \cite{kai2026cognidrive}.

\bibliographystyle{ACM-Reference-Format}
\bibliography{hri}

\clearpage

\appendix

\counterwithin{figure}{section}
\counterwithin{table}{section}
\counterwithin{equation}{section}

\section{Active Inference Theoretical Background}
\label{app:A}

Here we provide the formal mathematical formulations underpinning the Active Inference framework used in the main text.

\subsection{Perception as Inference: Minimizing Variational Free Energy}
Perception is modeled as a process of inferring the hidden causes of sensory signals by minimizing \textit{variational free energy (VFE)}. VFE provides a tractable upper bound on sensory surprise $-\log P(o)$. Mathematically, VFE reveals the core objective of Bayesian inference by relating the agent's belief to the true posterior:
\begin{equation}
    \label{eq:vfe_divergence_evidence}
    F(Q, o) = \underbrace{D_{KL}[Q(s) \| P(s \mid o)]}_{\text{Divergence}} \underbrace{- \log P(o)}_{\text{Surprise}}
\end{equation}
Here, $s$ represents the hidden environmental states, $o$ is the sensory observation, and $Q(s)$ is the agent's variational belief. Equation~\ref{eq:vfe_divergence_evidence} shows that for a given observation $o$, its surprise is constant. Therefore, minimizing VFE is equivalent to minimizing the Kullback-Leibler (KL) divergence between the agent's belief $Q(s)$ and the true posterior $P(s \mid o)$. This principle ensures that the agent's belief becomes the best possible approximation of the true posterior distribution over hidden states given the available data.

In our framework, VFE minimization governs belief dynamics. It compels belief updates in response to new evidence while ensuring belief persistence during non-informative periods (such as occlusions), as any arbitrary update without evidence would unjustifiably increase the KL divergence from the true posterior. This formally justifies our \textit{Conditional Reset} strategy, which prevents the belief in a latent threat from decaying without informative sensory data.

\subsection{Planning as Inference: Minimizing Expected Free Energy}
Action is defined as a process of inferring the best policy $\pi$ (sequence of actions) to execute. While perception minimizes surprise about current sensory data, action is selected to minimize expected surprise about the \textit{future}. This is achieved by selecting a policy $\pi$ that minimizes the \textit{Expected Free Energy (EFE)}, which elegantly unifies the trade-off between exploitation and exploration:
\begin{equation}
\label{eq:efe}
G(\pi) = -\underbrace{\mathbb{E}_{Q(o \mid \pi)}[\log P(o)]}_{\text{Pragmatic Value}} - \underbrace{\mathbb{E}_{Q(o \mid \pi)}\left[D_{K L}[Q(s \mid o, \pi) \| Q(s \mid \pi)]\right]}_{\text{Epistemic Value}}
\end{equation}
Here, \textit{pragmatic value} drives the agent towards preferred or goal-directed outcomes, encoded as a prior distribution over observations $P(o)$. In the context of pedestrian interaction, this motivates the vehicle to move forward while ensuring safety (such as avoiding a collision).
\textit{Epistemic value} encourages actions that yield the most informative future observations, such as peeking around an occlusion to reduce belief ambiguity about a potential pedestrian.

\paragraph{Simplification for Deterministic Models}
In the specific context of this work, we employ a deterministic observation model. Under this assumption, the epistemic value in Eq.~\ref{eq:efe} can be simplified to the entropy of predicted observations. The single-step EFE then takes the following form used to derive our particle approximation:
\begin{equation}
\label{eq:efe_simplified_theoretical}
G(\pi) = -\underbrace{\mathbb{E}_{Q(o \mid \pi)}[\log P(o)]}_{\text{Pragmatic Value}} -\underbrace{H[Q(o \mid \pi)]}_{\text{Epistemic Value}}
\end{equation}
This formulation highlights that the agent seeks to maximize the expected log-likelihood of preferred observations while simultaneously maximizing the entropy of the predicted observations (seeking information).

\section{Simulation Environment Details}
\label{app:B}

This appendix provides the specific parameter settings and distributions used in our simulation environment to reproduce the experimental results.

\subsection{System Parameters}
\subparagraph{\textbf{Ego-Vehicle}} The ego-vehicle has dimensions of length $L_e = 4.5$ m and width $W_e = 2.0$ m. It is initialized at state $x_{e,0} = (-25, 0)$ m, $v_{e,0} = (10, 0)$ m/s, and $a_{e,0} = (0, 0)$ m/s$^2$.
\subparagraph{\textbf{Static Environment}} The occluding object $\mathcal{M}_\text{occ}$ has dimensions $L_\text{occ} = 10$ m and $W_\text{occ} = 4$ m, centered at $x_\text{occ} = (5, -4)$ m. A collision is registered if the Euclidean distance between the ego-vehicle and the pedestrian is less than 1.0 m. The simulation time step is 0.1 seconds with a maximum duration of 150 steps.

\subsection{Pedestrian Latent Parameters and Behavioral Modes}

The latent behavioral parameters $\theta_p$ are categorized into common physical parameters shared across all modes and mode-specific parameters that govern distinct behavioral logic.

\subsubsection{Common Parameters}
All pedestrian agents share a set of fundamental kinematic parameters: the initial emergence position ($x_{p,0}$), maximum velocity capability ($v_{p, \text{max}}$), maximum acceleration capability ($a_{p, \text{max}}$), and the motion activation distance relative to the ego-vehicle ($d_\text{act}$). 

While $d_\text{act}$ is a common parameter type, its sampling distribution varies significantly across different behavioral modes to reflect different reaction sensitivities. The physical limits are sampled from uniform distributions: $v_{p, \text{max}} \sim \mathcal{U}(4.0, 8.0)$ m/s and $a_{p, \text{max}} \sim \mathcal{U}(4.0, 8.0)$ m/s$^2$. 

The initial position $x_{p,0}$ follows a truncated 2D Gaussian distribution $\mathcal{N}(\mu_x, \Sigma_x)$ with mean $\mu_x = (12, -4)$ m and diagonal standard deviation $\sigma_x = (\frac{2}{3}, \frac{2}{3})$ m. To strictly ensure the pedestrian is initially occluded, this distribution is clipped within the spatial boundaries defined by the obstacle.

\subsubsection{Behavioral Modes and Specific Parameters}
We define five distinct behavioral modes. For each mode, we specify the sampling range for the activation distance $d_\text{act}$ and define the mode-specific parameters ($\theta_{p, \text{spec}}$) as follows:

\begin{itemize}[leftmargin=14pt]
    \item \textbf{Hesitant Pedestrian:} This agent represents indecisive behavior, with an activation distance $d_\text{act} \sim \mathcal{U}(15, 25)$ m. Its specific behavior is governed by alternating hesitation and movement intervals. The hesitation duration $t_\text{h} \sim \mathcal{U}(0, 1.0)$ s and the movement duration $t_\text{m} \sim \mathcal{U}(1.5, 2.5)$ s are sampled to control the frequency of these state transitions.

    \item \textbf{Deceptive Accelerating Pedestrian:} Simulating an agent that starts slowly but rushes upon vehicle approach, this mode uses $d_\text{act} \sim \mathcal{U}(15, 25)$ m. The agent maintains a low initial velocity $v_{p, \text{slow}} \sim \mathcal{U}(1.5, 4.0)$ m/s until the longitudinal distance to the ego-vehicle drops below a trigger threshold $d_{\text{trig, acc}} \sim \mathcal{U}(4.0, 8.0)$ m, at which point it accelerates to $v_{p, \text{max}}$.

    \item \textbf{Turning-Back Pedestrian:} This agent enters the lane but retreats when the vehicle is within a specific proximity. The activation distance is $d_\text{act} \sim \mathcal{U}(15, 25)$ m. The specific parameters include the retreat trigger distance $d_{\text{trig, ret}} \sim \mathcal{U}(4.0, 8.0)$ m (relative to the vehicle) and the lateral turning point $x_{\text{lat, turn}} \sim \mathcal{U}(2.0, 6.0)$ m, which determines the maximum lateral intrusion into the road before turning back.

    \item \textbf{Sudden Stop Pedestrian:} Representing an agent freezing in the roadway, this mode is initialized with a tighter activation distance of $d_\text{act} \sim \mathcal{U}(15, 20)$ m. The specific parameter is the lateral stop position $x_{\text{lat, stop}}$, sampled from $\mathcal{U}(0, 4.0)$ m, where the pedestrian comes to a complete halt.

    \item \textbf{Sudden Appearance Pedestrian:} Simulating a ``ghost probe'' scenario with minimal reaction time, this mode reduces the activation distance to $d_\text{act} \sim \mathcal{U}(10, 15)$ m. It relies primarily on this short distance and high initial acceleration to create urgency, without additional mode-specific latent parameters.
\end{itemize}

\section{Details of Ablation Studies}
\label{app:C}

This section presents a granular breakdown of performance metrics across the five distinct pedestrian behavioral modes defined in Appendix~\ref{app:B}. Tables~\ref{tab:ablation_feint_and_retreat} through \ref{tab:ablation_hesitation} detail the impact of varying the Initial Presence Belief ($B_0(z_p)$) and Hypothesis Injection Ratio ($\rho_H$) on safety and efficiency for the Turning-Back, Deceptive Accelerating, Sudden Stop, Sudden Appearance, and Hesitant scenarios, respectively.

\begin{table*}
  \caption{Influence of Initial Presence Belief ($B_0(z_p)$) and Hypothesis Injection Ratio ($\rho_{H}$) (Turning-Back Pedestrian)} 
  \label{tab:ablation_feint_and_retreat}
  
  \begin{tabular}{c c c c c c c c}
  \toprule
   $\rho_H$ & $B_0(z_p)$ & \textbf{Pass Rate (PR)} $\uparrow$  & \textbf{Collision Rate (CR)} $\downarrow$ & \textbf{Pass Time (PT)} $\downarrow$ & \textbf{Min. Distance (MD)} $\uparrow$ & \textbf{Min. TTC (TTC)} $\uparrow$ \\
  \midrule
  0.0 & 0.8 & $0.783 \pm 0.038$ & $0.217 \pm 0.038$ & $ \textbf{7.685} \pm \textbf{0.056}$ & $\textbf{3.806} \pm \textbf{0.125}$ & $\textbf{1.798} \pm \textbf{0.113}$ \\
  \addlinespace
  0.1 & 0.8 & $0.858 \pm 0.032$ & $0.142 \pm 0.032$ & $7.700 \pm 0.061$ & $3.762 \pm 0.123$ & $1.659 \pm 0.105$ \\
  \addlinespace
  0.3 & 0.8 & $\textbf{0.925} \pm \textbf{0.024}$ & $\textbf{0.075} \pm \textbf{0.024}$ & $8.172 \pm 0.076$ & $3.651 \pm 0.093$ & $1.519 \pm 0.081$ \\
  \addlinespace
  0.8 & 0.8 & $0.908 \pm 0.026$ & $0.092 \pm 0.026$ & $8.231 \pm 0.101$ & $3.659 \pm 0.067$ & $1.216 \pm 0.043$ \\
  \midrule
  0.3 & 0.0 & $0.433 \pm 0.045$ & $0.567 \pm 0.045$ & $\textbf{6.452} \pm \textbf{0.013}$ & $\textbf{3.687} \pm \textbf{0.140}$ & $1.173 \pm 0.080$ \\
  \addlinespace
  0.3 & 0.2 & $0.783 \pm 0.038$ & $0.217 \pm 0.038$ & $7.040 \pm 0.057$ & $3.537 \pm 0.097$ & $1.107 \pm 0.056$ \\
  \addlinespace
  0.3 & 0.4 & $0.833 \pm 0.034$ & $0.167 \pm 0.034$ & $7.472 \pm 0.065$ & $3.675 \pm 0.101$ & $1.249 \pm 0.057$ \\
  \addlinespace
  0.3 & 0.8 & $\textbf{0.925} \pm \textbf{0.024}$ & $\textbf{0.075} \pm \textbf{0.024}$ & $8.172 \pm 0.076$ & $3.651 \pm 0.093$ & $\textbf{1.519} \pm \textbf{0.081}$ \\
  \bottomrule
  \end{tabular}
\end{table*}

\begin{table*}
  \caption{Influence of Initial Presence Belief ($B_0(z_p)$) and Hypothesis Injection Ratio ($\rho_{H}$) (Deceptive Accelerating Pedestrian)} 
  \label{tab:ablation_hidden_acceleration}
  
  \begin{tabular}{c c c c c c c c}
  \toprule
   $\rho_H$ & $B_0(z_p)$ & \textbf{Pass Rate (PR)} $\uparrow$  & \textbf{Collision Rate (CR)} $\downarrow$ & \textbf{Pass Time (PT)} $\downarrow$ & \textbf{Min. Distance (MD)} $\uparrow$ & \textbf{Min. TTC (TTC)} $\uparrow$ \\
  \midrule
  0.0 & 0.8 & $0.808 \pm 0.036$ & $0.192 \pm 0.036$ & $\textbf{7.721} \pm \textbf{0.052}$ & $3.723 \pm 0.072$ & $1.799 \pm 0.074$ \\
  \addlinespace
  0.1 & 0.8 & $0.758 \pm 0.039$ & $0.242 \pm 0.039$ & $7.796 \pm 0.058$ & $4.144 \pm 0.083$ & $\textbf{1.838} \pm \textbf{0.066}$ \\
  \addlinespace
  0.3 & 0.8 & $0.850 \pm 0.033$ & $0.150 \pm 0.033$ & $8.692 \pm 0.125$ & $\textbf{4.249} \pm \textbf{0.093}$ & $1.637 \pm 0.065$ \\
  \addlinespace
  0.8 & 0.8 & $\textbf{0.983} \pm \textbf{0.014}$ & $\textbf{0.017} \pm \textbf{0.012}$ & $10.431 \pm 0.139$ & $3.278 \pm 0.040$ & $0.872 \pm 0.031$ \\
  \midrule
  0.3 & 0.0 & $0.167 \pm 0.034$ & $0.833 \pm 0.034$ & $\textbf{6.435} \pm \textbf{0.018}$ & $3.141 \pm 0.103$ & $0.954 \pm 0.153$ \\
  \addlinespace
  0.3 & 0.2 & $0.558 \pm 0.045$ & $0.442 \pm 0.045$ & $7.057 \pm 0.080$ & $3.729 \pm 0.071$ & $1.100 \pm 0.048$ \\
  \addlinespace
  0.3 & 0.4 & $0.767 \pm 0.039$ & $0.233 \pm 0.039$ & $7.748 \pm 0.122$ & $3.943 \pm 0.083$ & $1.225 \pm 0.045$ \\
  \addlinespace
  0.3 & 0.8 & $\textbf{0.850} \pm \textbf{0.033}$ & $\textbf{0.150} \pm \textbf{0.033}$ & $8.692 \pm 0.125$ & $\textbf{4.249} \pm \textbf{0.093}$ & $\textbf{1.637} \pm \textbf{0.065}$ \\
  \bottomrule
  \end{tabular}
\end{table*}

\begin{table*}
  \caption{Influence of Initial Presence Belief ($B_0(z_p)$) and Hypothesis Injection Ratio ($\rho_{H}$) (Sudden Stop Pedestrian)} 
  \label{tab:ablation_sudden_stopping}
  
  \begin{tabular}{c c c c c c c c}
  \toprule
   $\rho_H$ & $B_0(z_p)$ & \textbf{Pass Rate (PR)} $\uparrow$  & \textbf{Collision Rate (CR)} $\downarrow$ & \textbf{Pass Time (PT)} $\downarrow$ & \textbf{Min. Distance (MD)} $\uparrow$ & \textbf{Min. TTC (TTC)} $\uparrow$ \\
  \midrule
  0.0 & 0.8 & $0.933 \pm 0.023$ & $0.067 \pm 0.023$ & $ 8.248 \pm 0.049$ & $2.713 \pm 0.030$ & $0.939 \pm 0.092$ \\
  \addlinespace
  0.1 & 0.8 & $\textbf{1.000} \pm \textbf{0.000}$ & $\textbf{0.000} \pm \textbf{0.000}$ & $\textbf{8.116} \pm \textbf{0.038}$ & $2.961 \pm 0.030$ & $\textbf{1.212} \pm \textbf{0.066}$ \\
  \addlinespace
  0.3 & 0.8 & $0.992 \pm 0.008$ & $0.008 \pm 0.008$ & $8.336 \pm 0.051$ & $\textbf{3.267} \pm \textbf{0.023}$ & $1.056 \pm 0.044$ \\
  \addlinespace
  0.8 & 0.8 & $\textbf{1.000} \pm \textbf{0.000}$ & $\textbf{0.000} \pm \textbf{0.000}$ & $8.363 \pm 0.058$ & $3.241 \pm 0.017$ & $0.828 \pm 0.015$ \\
  \midrule
  0.3 & 0.0 & $0.042 \pm 0.018$ & $0.958 \pm 0.018$ & $\textbf{6.420} \pm \textbf{0.033}$ & $2.613 \pm 0.143$ & $0.490 \pm 0.244$ \\
  \addlinespace
  0.3 & 0.2 & $0.692 \pm 0.042$ & $0.308 \pm 0.042$ & $7.343 \pm 0.062$ & $3.000 \pm 0.036$ & $0.675 \pm 0.035$ \\
  \addlinespace
  0.3 & 0.4 & $0.917 \pm 0.025$ & $0.083 \pm 0.025$ & $7.972 \pm 0.058$ & $3.243 \pm 0.028$ & $0.896 \pm 0.029$ \\
  \addlinespace
  0.3 & 0.8 & $\textbf{0.992} \pm \textbf{0.008}$ & $\textbf{0.008} \pm \textbf{0.008}$ & $8.336 \pm 0.051$ & $\textbf{3.267} \pm \textbf{0.023}$ & $\textbf{1.056} \pm \textbf{0.044}$ \\
  \bottomrule
  \end{tabular}
\end{table*}

\begin{table*}
  \caption{Influence of Initial Presence Belief ($B_0(z_p)$) and Hypothesis Injection Ratio ($\rho_{H}$) (Sudden Appearance Pedestrian)} 
  \label{tab:ablation_sudden_appearance}
  
  \begin{tabular}{c c c c c c c c}
  \toprule
   $\rho_H$ & $B_0(z_p)$ & \textbf{Pass Rate (PR)} $\uparrow$  & \textbf{Collision Rate (CR)} $\downarrow$ & \textbf{Pass Time (PT)} $\downarrow$ & \textbf{Min. Distance (MD)} $\uparrow$ & \textbf{Min. TTC (TTC)} $\uparrow$ \\
  \midrule
  0.0 & 0.8 & $\textbf{1.000} \pm \textbf{0.000}$ & $\textbf{0.000} \pm \textbf{0.000}$ & $ \textbf{8.586} \pm \textbf{0.033}$ & $3.213 \pm 0.042$ & $\textbf{1.636} \pm \textbf{0.063}$ \\
  \addlinespace
  0.1 & 0.8 & $\textbf{1.000} \pm \textbf{0.000}$ & $\textbf{0.000} \pm \textbf{0.000}$ & $8.790 \pm 0.054$ & $3.399 \pm 0.042$ & $1.511 \pm 0.063$ \\
  \addlinespace
  0.3 & 0.8 & $\textbf{1.000} \pm \textbf{0.000}$ & $\textbf{0.000} \pm \textbf{0.000}$ & $9.832 \pm 0.129$ & $\textbf{3.661} \pm \textbf{0.051}$ & $1.475 \pm 0.058$ \\
  \addlinespace
  0.8 & 0.8 & $0.992 \pm 0.016$ & $0.008 \pm 0.008$ & $9.445 \pm 0.143$ & $3.107 \pm 0.022$ & $0.730 \pm 0.018$ \\
  \midrule
  0.3 & 0.0 & $0.000 \pm 0.000$ & $1.000 \pm 0.000$ & $-$ & $-$ & $-$ \\
  \addlinespace
  0.3 & 0.2 & $0.533 \pm 0.046$ & $0.467 \pm 0.046$ & $\textbf{7.461} \pm \textbf{0.103}$ & $3.216 \pm 0.058$ & $0.571 \pm 0.018$ \\
  \addlinespace
  0.3 & 0.4 & $0.892 \pm 0.028$ & $0.108 \pm 0.028$ & $8.450 \pm 0.114$ & $3.236 \pm 0.036$ & $0.802 \pm 0.029$ \\
  \addlinespace
  0.3 & 0.8 & $\textbf{1.000} \pm \textbf{0.000}$ & $\textbf{0.000} \pm \textbf{0.000}$ & $9.832 \pm 0.129$ & $\textbf{3.661} \pm \textbf{0.051}$ & $\textbf{1.475} \pm \textbf{0.058}$ \\
  \bottomrule
  \end{tabular}
\end{table*}

\begin{table*}
  \caption{Influence of Initial Presence Belief ($B_0(z_p)$) and Hypothesis Injection Ratio ($\rho_{H}$) (Hesitant Pedestrian)} 
  \label{tab:ablation_hesitation}
  
  \begin{tabular}{c c c c c c c c}
  \toprule
   $\rho_H$ & $B_0(z_p)$ & \textbf{Pass Rate (PR)} $\uparrow$  & \textbf{Collision Rate (CR)} $\downarrow$ & \textbf{Pass Time (PT)} $\downarrow$ & \textbf{Min. Distance (MD)} $\uparrow$ & \textbf{Min. TTC (TTC)} $\uparrow$ \\
  \midrule
  0.0 & 0.8 & $\textbf{1.000} \pm \textbf{0.000}$ & $\textbf{0.000} \pm \textbf{0.000}$ & $ \textbf{7.537} \pm \textbf{0.031}$ & $4.913 \pm 0.113$ & $\textbf{2.496} \pm \textbf{0.033}$ \\
  \addlinespace
  0.1 & 0.8 & $\textbf{1.000} \pm \textbf{0.000}$ & $\textbf{0.000} \pm \textbf{0.000}$ & $7.610 \pm 0.034$ & $\textbf{5.076} \pm \textbf{0.111}$ & $2.463 \pm 0.033$ \\
  \addlinespace
  0.3 & 0.8 & $0.967 \pm 0.016$ & $0.033 \pm 0.016$ & $8.186 \pm 0.094$ & $4.678 \pm 0.107$ & $2.039 \pm 0.055$ \\
  \addlinespace
  0.8 & 0.8 & $\textbf{1.000} \pm \textbf{0.000}$ & $\textbf{0.000} \pm \textbf{0.000}$ & $9.227 \pm 0.143$ & $3.568 \pm 0.078$ & $1.183 \pm 0.037$ \\
  \midrule
  0.3 & 0.0 & $0.608 \pm 0.045$ & $0.392 \pm 0.045$ & $\textbf{6.451} \pm \textbf{0.010}$ & $4.080 \pm 0.121$ & $1.377 \pm 0.046$ \\
  \addlinespace
  0.3 & 0.2 & $0.967 \pm 0.016$ & $0.033 \pm 0.016$ & $6.950 \pm 0.029$ & $4.114 \pm 0.096$ & $1.321 \pm 0.048$ \\
  \addlinespace
  0.3 & 0.4 & $\textbf{0.992} \pm \textbf{0.008}$ & $\textbf{0.008} \pm \textbf{0.008}$ & $7.329 \pm 0.044$ & $4.352 \pm 0.092$ & $1.576 \pm 0.046$ \\
  \addlinespace
  0.3 & 0.8 & $0.967 \pm 0.016$ & $0.033 \pm 0.016$ & $8.186 \pm 0.094$ & $\textbf{4.678} \pm \textbf{0.107}$ & $\textbf{2.039} \pm \textbf{0.055}$ \\
  \bottomrule
  \end{tabular}
\end{table*}

\section{Details of Comparison with Baseline Methods}
\label{app:D}

This section provides detailed quantitative comparisons between our method and baseline approaches across the five behavioral modes. Tables~\ref{tab:baseline_feint_and_retreat} through \ref{tab:baseline_hesitation} report the specific performance metrics for the Turning-Back, Deceptive Accelerating, Sudden Stop, Sudden Appearance, and Hesitant pedestrian scenarios, respectively.

\begin{table*}
  \caption{Quantitative Comparison with Baseline Methods (Turning-Back Pedestrian)}
  \label{tab:baseline_feint_and_retreat}
  \begin{tabular}{l c c c c c c }
  \toprule
   \textbf{Method} & \textbf{Pass Rate (PR)} $\uparrow$  & \textbf{Collision Rate (CR)} $\downarrow$ & \textbf{Pass Time (PT)} $\downarrow$ & \textbf{Min. Distance (MD)} $\uparrow$ & \textbf{Min. TTC (TTC)} $\uparrow$ \\
  \midrule
  Reactive & $0.325 \pm 0.043$ & $0.675 \pm 0.043$ & $ 6.441 \pm 0.060$ & $2.931 \pm 0.098$ & $0.932 \pm 0.113$ \\
  \addlinespace
  Rule-based & $0.792 \pm 0.037$ & $0.208 \pm 0.037$ & $ 7.674 \pm 0.074$ & $3.327 \pm 0.089$ & $0.855 \pm 0.058$ \\
  \addlinespace
  PPO-LSTM & $0.650 \pm 0.044$ & $0.350 \pm 0.044$ & $\textbf{4.172} \pm \textbf{0.005}$ & $\textbf{3.772} \pm \textbf{0.087}$ & $\textbf{8.493} \pm \textbf{0.400}$ \\
  \midrule
  Ours & $\textbf{0.925} \pm \textbf{0.024}$ & $\textbf{0.075} \pm \textbf{0.024}$ & $8.172 \pm 0.076$ & $3.651 \pm 0.093$ & $1.519 \pm 0.081$ \\
  \bottomrule
  \end{tabular}
\end{table*}

\begin{table*}
  \caption{Quantitative Comparison with Baseline Methods (Deceptive Accelerating Pedestrian)}
  \label{tab:baseline_hidden_acceleration}
  \begin{tabular}{l c c c c c c }
  \toprule
   \textbf{Method} & \textbf{Pass Rate (PR)} $\uparrow$  & \textbf{Collision Rate (CR)} $\downarrow$ & \textbf{Pass Time (PT)} $\downarrow$ & \textbf{Min. Distance (MD)} $\uparrow$ & \textbf{Min. TTC (TTC)} $\uparrow$ \\
  \midrule
  Reactive & $0.000 \pm 0.000$ & $1.000 \pm 0.000$ & $-$ & $-$ & $-$ \\
  \addlinespace
  Rule-based & $0.267 \pm 0.040$ & $0.733 \pm 0.040$ & $ 8.619 \pm 0.130$ & $2.557 \pm 0.034$ & $0.605 \pm 0.073$ \\
  \addlinespace
  PPO-LSTM & $0.733 \pm 0.040$ & $0.267 \pm 0.040$ & $\textbf{4.178} \pm \textbf{0.005}$ & $3.932 \pm 0.058$ & $\textbf{10.000} \pm \textbf{0.000}$ \\
  \midrule
  Ours & $\textbf{0.850} \pm \textbf{0.033}$ & $\textbf{0.150} \pm \textbf{0.033}$ & $8.692 \pm 0.125$ & $\textbf{4.249} \pm \textbf{0.093}$ & $1.637 \pm 0.065$ \\
  \bottomrule
  \end{tabular}
\end{table*}

\begin{table*}
  \caption{Quantitative Comparison with Baseline Methods (Sudden Stop Pedestrian)}
  \label{tab:baseline_sudden_stop}
  \begin{tabular}{l c c c c c c }
  \toprule
   \textbf{Method} & \textbf{Pass Rate (PR)} $\uparrow$  & \textbf{Collision Rate (CR)} $\downarrow$ & \textbf{Pass Time (PT)} $\downarrow$ & \textbf{Min. Distance (MD)} $\uparrow$ & \textbf{Min. TTC (TTC)} $\uparrow$ \\
  \midrule
  Reactive & $0.200 \pm 0.037$ & $0.800 \pm 0.037$ & $ 6.742 \pm 0.027$ & $2.361 \pm 0.027$ & $0.053 \pm 0.007$ \\
  \addlinespace
  Rule-based & $0.892 \pm 0.028$ & $0.108 \pm 0.028$ & $ 7.496 \pm 0.018$ & $2.530 \pm 0.007$ & $0.260 \pm 0.029$ \\
  \addlinespace
  PPO-LSTM & $0.833 \pm 0.034$ & $0.167 \pm 0.034$ & $\textbf{4.195} \pm \textbf{0.002}$ & $\textbf{3.616} \pm \textbf{0.049}$ & $\textbf{9.708} \pm \textbf{0.166}$ \\
  \midrule
  Ours & $\textbf{0.992} \pm \textbf{0.008}$ & $\textbf{0.008} \pm \textbf{0.008}$ & $8.336 \pm 0.051$ & $3.267 \pm 0.023$ & $1.056 \pm 0.044$ \\
  \bottomrule
  \end{tabular}
\end{table*}

\begin{table*}
  \caption{Quantitative Comparison with Baseline Methods (Sudden Appearance Pedestrian)}
  \label{tab:baseline_sudden_appearance}
  \begin{tabular}{l c c c c c c }
  \toprule
   \textbf{Method} & \textbf{Pass Rate (PR)} $\uparrow$  & \textbf{Collision Rate (CR)} $\downarrow$ & \textbf{Pass Time (PT)} $\downarrow$ & \textbf{Min. Distance (MD)} $\uparrow$ & \textbf{Min. TTC (TTC)} $\uparrow$ \\
  \midrule
  Reactive & $0.000 \pm 0.000$ & $1.000 \pm 0.000$ & $-$ & $-$ & $-$ \\
  \addlinespace
  Rule-based & $0.133 \pm 0.031$ & $0.867 \pm 0.031$ & $ 7.706 \pm 0.209$ & $2.771 \pm 0.118$ & $0.682 \pm 0.136$ \\
  \addlinespace
  PPO-LSTM & $\textbf{1.000} \pm \textbf{0.000}$ & $\textbf{0.000} \pm \textbf{0.000}$ & $\textbf{4.200} \pm \textbf{0.000}$ & $\textbf{4.947} \pm \textbf{0.048}$ & $\textbf{10.000} \pm \textbf{0.000}$ \\
  \midrule
  Ours & $\textbf{1.000} \pm \textbf{0.000}$ & $\textbf{0.000} \pm \textbf{0.000}$ & $9.832 \pm 0.129$ & $3.661 \pm 0.051$ & $1.475 \pm 0.058$ \\
  \bottomrule
  \end{tabular}
\end{table*}

\begin{table*}
  \caption{Quantitative Comparison with Baseline Methods (Hesitant Pedestrian)}
  \label{tab:baseline_hesitation}
  \begin{tabular}{l c c c c c c }
  \toprule
   \textbf{Method} & \textbf{Pass Rate (PR)} $\uparrow$  & \textbf{Collision Rate (CR)} $\downarrow$ & \textbf{Pass Time (PT)} $\downarrow$ & \textbf{Min. Distance (MD)} $\uparrow$ & \textbf{Min. TTC (TTC)} $\uparrow$ \\
  \midrule
  Reactive & $0.367 \pm 0.044$ & $0.633 \pm 0.044$ & $ 6.441 \pm 0.081$ & $3.531 \pm 0.148$ & $1.174 \pm 0.094$ \\
  \addlinespace
  Rule-based & $0.850 \pm 0.033$ & $0.150 \pm 0.033$ & $ 7.839 \pm 0.069$ & $3.262 \pm 0.103$ & $0.877 \pm 0.069$ \\
  \addlinespace
  PPO-LSTM & $0.408 \pm 0.045$ & $0.592 \pm 0.045$ & $\textbf{4.188} \pm \textbf{0.005}$ & $3.659 \pm 0.077$ & $\textbf{10.000} \pm \textbf{0.000}$ \\
  \midrule
  Ours & $\textbf{0.967} \pm \textbf{0.016}$ & $\textbf{0.033} \pm \textbf{0.016}$ & $8.186 \pm 0.094$ & $\textbf{4.678} \pm \textbf{0.107}$ & $2.039 \pm 0.055$ \\
  \bottomrule
  \end{tabular}
\end{table*}

\end{document}